\pdfoutput=1
\documentclass[10pt,twocolumn,letterpaper]{article}
\usepackage[pagenumbers]{cvpr}
\def\HideWritingNotes{1}
\usepackage{dblfloatfix}
\usepackage{comment}
\usepackage{multirow}
\usepackage{listings}
\usepackage{float}
\usepackage{makecell}
\newif\ifwritingnotes
\ifdefined\HideWritingNotes\writingnotesfalse\else\writingnotestrue\fi
\definecolor{writingorange}{RGB}{153,68,0}
\definecolor{evidencepurple}{RGB}{106,45,135}

\usepackage{array}
\usepackage{tabularx}
\usepackage{placeins}
\floatstyle{ruled}
\newfloat{algorithm}{tbp}{loa}
\floatname{algorithm}{Algorithm}

\floatstyle{plain}
\providecommand{\nolinenumbers}{}
\definecolor{cvprblue}{rgb}{0.21,0.49,0.74}
\usepackage[pagebackref,breaklinks,colorlinks,allcolors=cvprblue]{hyperref}
\def\confName{CVPR}
\def\confYear{2027}
\title{RLM: Learning Real-World Automotive Radar Perception through Language Supervision in Simulation}
\author{Pushkal Mishra$^{1}$ \qquad Kshitiz Bansal$^{2}$ \qquad Dinesh Bharadia$^{1}$\\[0.4em]
$^{1}$University of California San Diego \qquad $^{2}$Blue River Technology\\[0.2em]
{\tt\small \{pumishra, dbharadia\}@ucsd.edu, \ ksbansal@ucsd.edu}}

\begin{document}
\addtocontents{toc}{\protect\setcounter{tocdepth}{-10}}
\maketitle

\begin{abstract}
    Radar supports perception at long range and under adverse weather and low visibility, making it an important sensing modality for autonomous driving. However, radar learning remains fragmented across task-specific architectures and objectives, while limited real-world annotations hinder the development of transferable representations. We present Radar Language Model (RLM), a framework for pretraining radar representations through spatial language supervision in simulation. By describing vehicle counts and their spatial arrangement, language provides a common supervisory signal for tasks with different prediction outputs. We construct over 800,000 radar-caption pairs from 110 hours of diverse simulated driving. We also introduce a new contrastive objective that incorporates spatial scene similarity into radar-language alignment, encouraging representations that capture the structure expressed in these descriptions. The resulting features support scene captioning, vehicle segmentation and bounding-box detection after supervised adaptation to RADIATE dataset. Experiments demonstrate improved count-based F1, IoU, and Dice over vanilla CLIP, alongside the highest mean detection AP50 among the evaluated baselines. Detection gains persist under limited real-world annotation budgets, supporting spatial-language pretraining as a practical route to reusable radar representations.
\end{abstract}



\section{Introduction}
\label{sec:intro}

Radar is an important sensing modality for autonomous driving because it provides range and motion measurements in conditions that challenge optical perception, including poor visibility and adverse weather~\cite{radar-bw-3,radiate}. Turning these measurements into useful perception features, however, remains constrained by the availability of training supervision. Radar returns can be difficult to interpret and annotate, making large-scale real-world labeling expensive~\cite{bootstrap-radar}. At the same time, detection~\cite{rodnet,pointillism,radnet,radatron,radsegnet,racformer}, dense prediction~\cite{radar-segmentation-wsss,radar-road-segmentation}, and motion estimation~\cite{4d-scene-flow-radar} require different outputs and annotation formats. Learning representations that can be reused across these tasks is therefore particularly valuable when real-world labels are limited.

Existing work offers several routes toward this goal. Cross-modal supervision uses cameras, LiDAR, and other co-recorded signals to train radar perception models with less direct radar annotation~\cite{rodnet,4d-scene-flow-radar}. RADICaL combines radar-radar and radar-vision contrastive learning to improve object detection~\cite{bootstrap-radar}. The Generalizable Radar Transformer demonstrates the value of large-scale raw-radar learning and adaptation across spatial prediction tasks~\cite{radar-fm-single-chip}. These results establish the utility of reusable radar features, while also highlighting the importance of the data and supervision used to learn them. However, reducing manual annotation does not remove the cost of collecting real sensor data across diverse scenes and acquisition settings.

\begin{figure*}[t]
    \centering
    \includegraphics[width=0.95\textwidth]{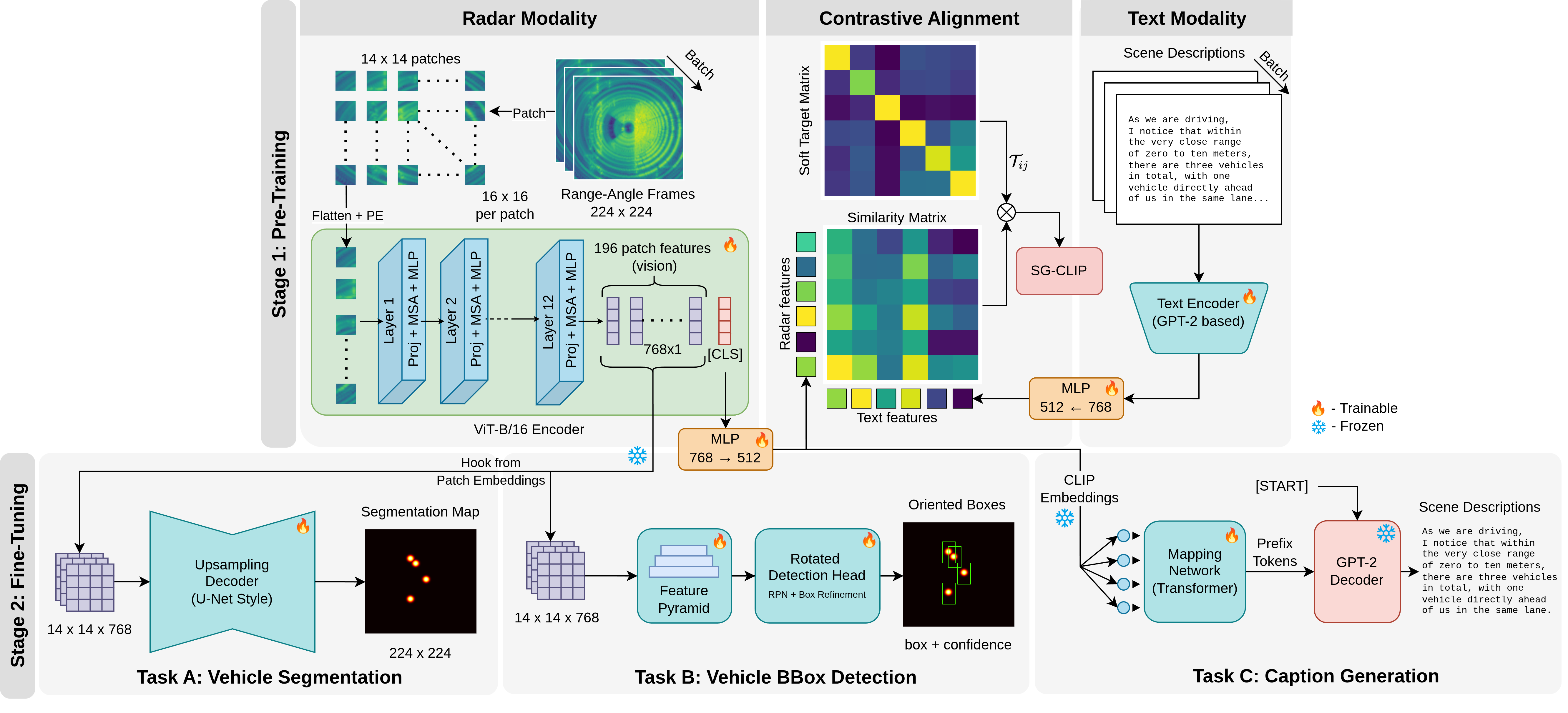}
    \caption{Overview of the Radar Language Model (RLM) framework. Radar range-angle heatmaps are encoded by a ViT-B/16 vision encoder, while structured spatial captions are processed by a causal Transformer-based text encoder. Both modalities are projected into a shared embedding space where SG-CLIP aligns representations based on continuous scene similarity. The frozen encoder is subsequently validated via generative captioning (from the CLS token), vehicle segmentation and bounding box detections (from patch tokens).}
    \label{fig:architecture}
\end{figure*}

Simulation offers a practical source of such supervision. Driving simulators provide controllable scenes and actor geometry, and radar simulation models make corresponding sensor observations available~\cite{carla-sim,shenron,c-shenron,c-shenron-demo}. Their value for perception is supported by synthetic-to-real transfer results in radar object detection~\cite{radsimreal,sim2radar-vlm}. These studies supervise detection outputs, while simulation also exposes the underlying spatial organization of a scene. This motivates a representation-learning question: can that information provide a shared pretraining signal whose benefits carry over to different real-world radar tasks?


Language offers a \textit{universal label space} for this shared supervision. A description such as ``three vehicles in the right adjacent lane between 10 and 20\,m ahead'' expresses count, distance, and relative arrangement without prescribing a bounding-box or dense-map output. In vision, CLIP and BLIP-2 demonstrate the value of aligning visual features with language~\cite{clip,blip}, while language-guided detection~\cite{glip} and segmentation~\cite{clip-s4,mask-clip} extend this connection to spatial prediction. Radar-language models have explored this connection for specific applications, including text-conditioned object grounding~\cite{talk2radar} and human-motion understanding using synthetic radar-text pairs~\cite{radarllm-aaai2026}. For automotive perception, vehicle counts and spatial arrangement are relevant to both scene description and localization, suggesting that their expression in language can provide shared supervision across these different outputs. Such pretraining may be particularly useful when real-world labels are limited, allowing task-specific learning to build on spatial information acquired in simulation.

We introduce \textbf{Radar Language Model (RLM)} (Figure~\ref{fig:architecture}), a framework for learning transferable radar representations through structured spatial language supervision. Using CARLA with a radar simulation model, we construct over 800,000 radar-caption pairs spanning 110 hours of diverse simulated driving, describing vehicle distributions in ego-centric spatial regions. To distinguish partially similar scenes from unrelated ones, our \textbf{Spatially-Grounded CLIP (SG-CLIP)} objective guides radar-language alignment using similarity between these distributions. We evaluate the learned representations through scene captioning in simulation and supervised adaptation to vehicle segmentation and oriented bounding-box detection on RADIATE~\cite{radiate}. SG-CLIP achieves the highest mean detection AP50 among the evaluated baselines under full and limited annotation budgets, with complementary gains in captioning and vehicle-center prediction.

\noindent Our contributions are therefore threefold:
\begin{itemize}
    \item \textbf{Scalable supervision from simulation:} We construct a large radar-language dataset that expresses vehicle distributions through a shared spatial description, providing pretraining supervision before real-world task annotation.
    \item \textbf{Spatially-Grounded CLIP:} We introduce a contrastive pretraining objective that incorporates explicit scene similarity derived from spatial vehicle counts into radar-language alignment.
    \item \textbf{Transfer to real-world radar perception:} We evaluate the resulting approach through supervised detection and segmentation prediction on RADIATE, including full- and limited-label comparisons, complemented by scene-captioning evaluation in simulation.
\end{itemize}

\section{Related Works}
\label{sec:rel-works}

\textbf{Vision-language models and contrastive learning:}
CLIP~\cite{clip} trains image and text encoders to align paired samples in a shared embedding space through contrastive learning.
This positions language as a universal supervision signal: a common semantic interface supporting large-scale representation learning~\cite{align-google,florence}, grounded and dense prediction~\cite{glip,region-clip,mask-clip,clip-s4,vlm-for-segmentation-review}, and generation~\cite{blip,flamingo,coca}.
Complementary foundation models support promptable image and video segmentation~\cite{sam,sam-2}.
SpatialCLIP emphasizes spatially discriminative descriptions~\cite{spatial-clip}, while contrastive approaches incorporate similarity and ranking~\cite{soft-clip-loss,rank-clip}, class or prototype structure~\cite{supervised-clip-loss,proto-nce}, sigmoid losses~\cite{sigclip}, and joint self-supervision~\cite{slip}.
\textit{Why use CLIP?} Its cross-modal matching can exhibit bag-of-words behavior~\cite{clip-bag-of-words}.
We design our spatial captions around this property, encoding scene semantics as structured distributions of vehicle counts across ego-centric regions.

\textbf{Radar perception for autonomous driving:}
Radar perception spans object detection~\cite{rodnet,radnet,radatron,trodnet2023}, dense segmentation~\cite{radar-road-segmentation,radar-segmentation-wsss}, and scene-flow estimation~\cite{4d-scene-flow-radar}, with different output representations and supervision requirements.
Sensor configurations range from multi-radar bounding-box estimation~\cite{pointillism} to radar-camera fusion~\cite{radsegnet,racformer}, while K-Radar and RADIATE support evaluation across driving and weather conditions~\cite{k-radar,radiate}.
Cross-modal and weak supervision reduce direct radar annotation requirements~\cite{rodnet,4d-scene-flow-radar,radar-road-segmentation}, but the diversity of prediction targets and architectures motivates reusable pretraining.
TempoRadar and SIRA exploit multiple radar frames~\cite{temporadar,sira}, whereas our evaluation isolates single-frame representation learning; their reported scores therefore do not provide a directly matched comparison.

\textbf{Radar representation learning:}
Radar-language approaches address human activity and motion~\cite{mmclip,radarllm-aaai2026}, text-conditioned object grounding~\cite{talk2radar,talk2pc}, and scene description and reasoning~\cite{weather-robust-radar,radar4d-vlm}.
Broader representation-learning efforts include GRT's adaptation of raw-radar features across tasks and acquisition settings~\cite{radar-fm-single-chip} and RADICaL's radar-radar and radar-vision contrastive pretraining~\cite{bootstrap-radar}.
RSLM~\cite{radar-clip-captioning} uses paired real radar--camera observations to align radar with vision--language embeddings, supporting retrieval, detection, and free-space segmentation. Our approach instead learns from simulated radar--caption pairs, with SG-CLIP deriving alignment targets from similarity between spatial vehicle-count distributions.

\textbf{Simulation for radar learning:}
RadSimReal~\cite{radsimreal} combines simulated reflections with a measured radar point-spread function, demonstrating that detectors trained entirely on synthetic radar images can transfer to real datasets.
Sim2Radar~\cite{sim2radar-vlm} combines VLM-guided material reconstruction with radar simulation, demonstrating encoder pretraining for real-world detection.
While these works demonstrate synthetic-to-real transfer for detection, our work presents a practical method for constructing spatial-language supervision and shows that pretraining with this supervision in simulation improves performance across multiple real-world automotive radar tasks.

\section{Simulated Radar-Language Dataset}
\label{sec:dataset}

\begin{figure}[t]
    \centering
    \includegraphics[width=\columnwidth]{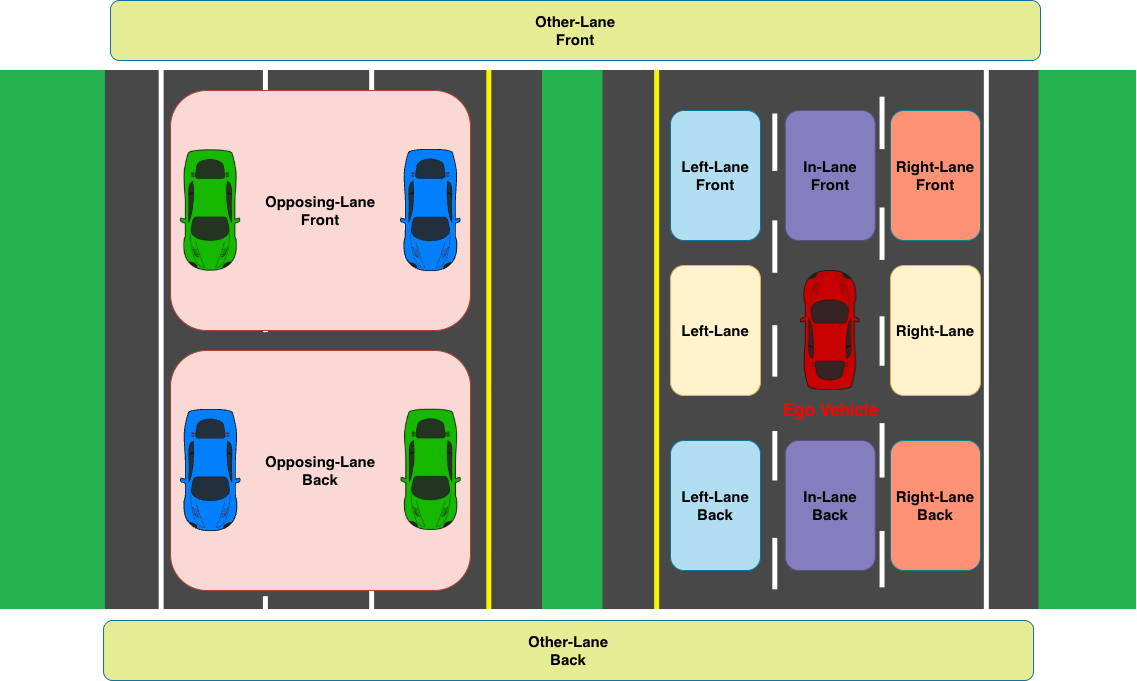}
    \caption{Lane distribution visualization showing the twelve lane-relative angular sectors used for spatial encoding of vehicles.}
    \label{fig:lane_distribution}
    \label{fig:dataset-division}
\end{figure}

\begin{figure*}[t]
    \centering
    \begin{subfigure}[t]{0.32\textwidth}
        \centering
        \includegraphics[width=\linewidth]{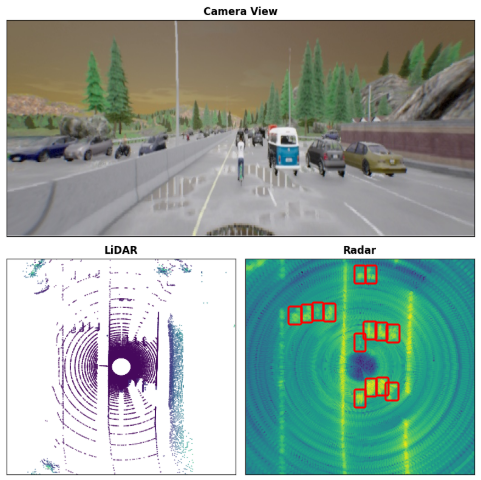}
        \caption{Recorded sensor views.}
        \label{fig:sensor_views}
    \end{subfigure}\hfill
    \begin{subfigure}[t]{0.32\textwidth}
        \centering
        \includegraphics[width=\linewidth]{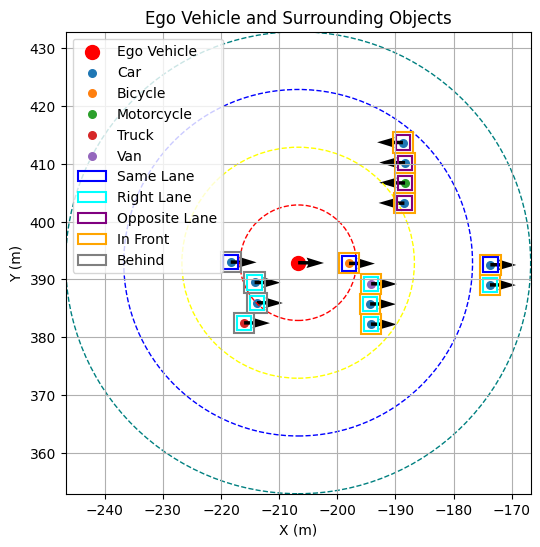}
        \caption{Extracted actor geometry.}
        \label{fig:surrounding}
    \end{subfigure}\hfill
    \begin{subfigure}[t]{0.32\textwidth}
        \centering
        \includegraphics[width=\linewidth]{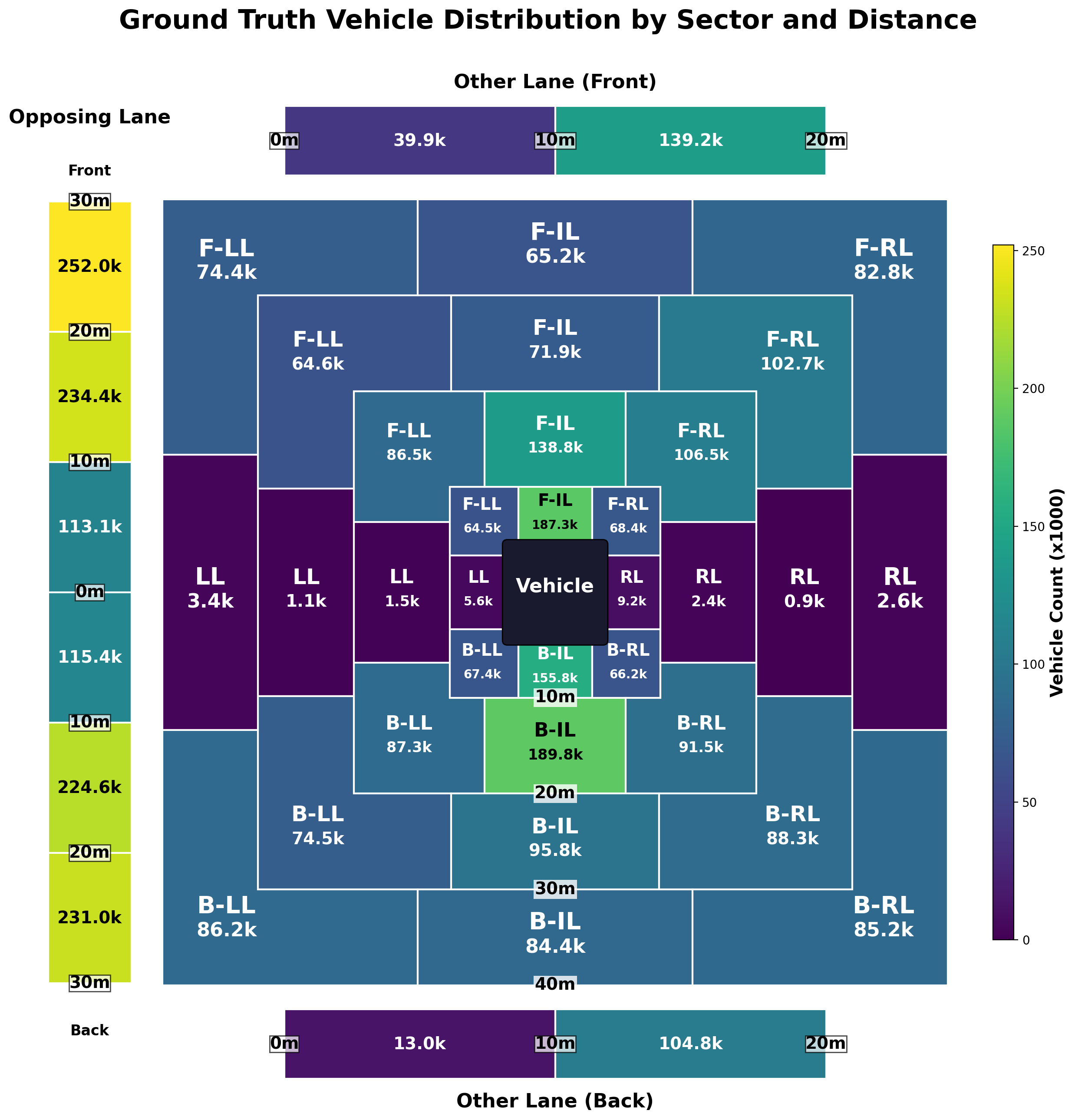}
        \caption{Vehicle distribution by relation.}
        \label{fig:lane-dist}
    \end{subfigure}
    \caption{Simulation data and spatial annotations. The collection example combines camera, LiDAR and radar views with the extracted vehicle positions and headings. Range rings are spaced at 10\,m. The distribution panel summarizes the corpus using the ego-relative relation categories (F/B: front/back; LL/IL/RL: left/in/right lane). We only use radar for our tasks.}
    \label{fig:dataset_collection}
\end{figure*}

Contrastive pretraining requires large-scale data to learn transferable representations~\cite{clip}, but collecting and annotating real-world radar at this scale is costly. We use CARLA~\cite{carla-sim} to obtain radar observations paired with spatial descriptions derived from simulator actor states. Since CARLA's radar sensor is highly simplified, we use an open-source implementation~\cite{c-shenron,c-shenron-demo,shenron}, which models the propagation behavior of automotive radars very accurately. We use the TI radar sensor configuration reported in C-Shenron~\cite{c-shenron,c-shenron-demo} for constructing our dataset. CARLA being a simulator, we can extract ground truth states of each vehicles and construct our captions using them.

\noindent\textbf{Natural language captions:}
The key challenge in constructing effective radar-text pairs is to describe \textit{where} vehicles are located relative to the ego vehicle, rather than merely stating their presence. To address this, we develop a structured encoding framework that partitions the radar scene into distance bins and angular sectors, grounding the descriptions in the spatial organization of the radar scene.

We divide the range from 0 to 40\,m into four equal bins and categorize vehicles into twelve spatial relations using their positions and headings relative to the ego vehicle (Figure~\ref{fig:lane_distribution}). This ego-centric spatial encoding provides the foundation for generating structured scene descriptions. Each scene's vehicle counts are stored in a structured JSON record. Figures~\ref{fig:sensor_views} and~\ref{fig:surrounding} show an example scene and its extracted actor geometry; below we show the corresponding vehicle distribution:

\begin{lstlisting}[float=htbp,basicstyle=\scriptsize\ttfamily,frame=single,breaklines=true,columns=fullflexible,caption={},label={lst:json-caption}]
{"0-10m": {"total_vehicles": 3,
           "in_lane_front_side": 1,
           "right_lane_back_side": 2},
 "10-20m": {"total_vehicles": 5,
            "in_lane_back_side": 1,
            "right_lane_front_side": 3,
            "right_lane_back_side": 1},
 "20-30m": {"total_vehicles": 4,
            "opposing_lane_front": 4},
 "30-40m": {"total_vehicles": 2,
            "in_lane_front_side": 1,
            "right_lane_front_side": 1}}
\end{lstlisting}

From these records, we construct both template-based and LLM-generated descriptions of the same vehicle distributions, using a mixture of the two for radar-language pretraining. The structured captions express counts and spatial relations consistently, while LLM-generated variants provide diversity in phrasing. During training, we randomly sample one of the captions for each radar frame. A corresponding natural-language description is:

\begin{small}
    \noindent\textit{``Within zero to ten meters, there are three vehicles: one directly ahead in the same lane and two in the right adjacent lane behind. Between ten and twenty meters, there are five vehicles: one behind in the same lane, three in the right adjacent lane ahead, and one in the right adjacent lane behind. Between twenty and thirty meters, four vehicles are in the opposing lane ahead. Between thirty and forty meters, there are two vehicles: one ahead in the same lane and one in the right adjacent lane ahead.''}
    \par
\end{small}

\noindent\textbf{Data collection and statistics:}
Collection spans 110 hours of simulated driving across multiple CARLA towns and yields 879{,}383 radar frames with paired spatial descriptions. We cover urban, rural, highway and intersection scenarios under varying traffic densities, from empty scenes to dense surrounding traffic. Figure~\ref{fig:lane-dist} summarizes the spatial distribution of vehicles. Further collection and implementation details are provided in the supplementary materials. This dataset is, to our knowledge, the first large-scale radar-language dataset, and plan on releasing it to facilitate future research in radar-language learning.

\section{RLM: Architecture and Approach}
\label{sec:approach}

RLM learns radar representations through spatial language supervision in simulation (Figure~\ref{fig:architecture}). A radar encoder and a text encoder map paired observations and descriptions into a shared embedding space, where SG-CLIP aligns them according to the similarity of their vehicle distributions. Scene-level contrastive alignment alone, however, does not establish spatial grounding across the representation.

We adapt the pretrained radar backbone through independently trained task heads. Captioning examines whether the global scene embedding encodes vehicle distributions across distance bins and spatial relations in the \textit{global CLS token}, while vehicle segmentation and bounding-box detection use patch features to localize vehicles from \textit{patch-level features}. These complementary outputs connect scene-level spatial semantics with local prediction, and real-world adaptation tests the utility of simulation pretraining.

\subsection{Pre-training Stage}
\label{sec:pretraining}

We adopt the pretrained ViT-B/16 image encoder from CLIP~\cite{clip,vit}, using its visual features as the initialization for radar-language pretraining. Each $224\times224$ radar range-angle map produces a global CLS token and 196 patch tokens arranged on a $14\times14$ spatial grid. The CLS token provides the scene representation for contrastive learning, while the patch tokens retain spatially organized features for downstream perception.

For language, we use a causal Transformer~\cite{transformers} following CLIP's text architecture. Our captions enumerate vehicle distributions across spatial regions and frequently exceed CLIP's 77-token context, so we extend the context window to 400 tokens and train the text encoder from scratch. Linear projections map the global radar feature and end-of-text representation into a shared 512-dimensional space, yielding embeddings $\mathbf{z}_r$ and $\mathbf{z}_t$. Both encoders are optimized during radar-language pretraining.

\subsection{Spatially-Grounded Contrastive Learning}
\label{sec:spatial-clip-loss}

Standard CLIP~\cite{clip} treats the paired image and caption as positive and all other pairs in a batch as negative. Radar scenes, however, can share similar vehicle distributions. For example, two otherwise identical scenes containing three versus two vehicles in the left lane ahead are more similar than scenes containing three versus none. SG-CLIP captures these relationships through soft contrastive targets derived from spatial count distributions, allowing partially similar scenes to contribute to representation learning.

Let $\mathbf{v}_i$ denote the encoded vehicle-count vector for scene $i$. Its entries $v_i^{(b,s)}$ correspond to range bin $b$ and relation $s$. We compute
\begin{equation*}
    d_{ij}=\sum_b\sum_s\left|v_i^{(b,s)}-v_j^{(b,s)}\right|
    \label{eq:kernel}
\end{equation*}
which represents the total count discrepancy across all distance bins for scene $i$ and scene $j$. The normalized soft target matrix is computed as
\begin{equation*}
    T_{ij} = \frac{\exp(-\alpha d_{ij}^{2})}{\sum_k \exp(-\alpha d_{ik}^{2})}
    \label{eq:soft-target}
\end{equation*}
The parameter $\alpha$ controls how quickly similarity decreases with count discrepancy: larger values concentrate supervision on matching vehicle distributions, while smaller values give more weight to partially similar scenes encouraging the model to learn from fine-grained differences. Row normalization produces a probability distribution over the captions in each batch.

Given radar embeddings $\mathbf{z}_r^i$ and text embeddings $\mathbf{z}_t^j$, we compute pairwise cosine similarities:
\begin{equation*}
    S_{ij} = \frac{\mathbf{z}_r^i \cdot \mathbf{z}_t^j}{\|\mathbf{z}_r^i\| \|\mathbf{z}_t^j\|}
\end{equation*}
and the radar-to-text objective is
\begin{equation*}
    \mathcal{L}_{r\rightarrow t} = -\frac{1}{N} \sum_i \sum_j T_{ij}\log\frac{\exp(S_{ij}/\tau)}{\sum_k\exp(S_{ik}/\tau)}
    \label{eq:sgclip}
\end{equation*}
where $N$ is the contrastive batch size and $\tau$ is the temperature. We apply the same scene-similarity target construction in the reverse direction and minimize the average of the two cross-entropies. This formulation provides two key advantages over standard CLIP:
\begin{itemize}
    \item \textbf{Fine-grained learning:} The model avoids harsh penalties for placing similar scenes close together in embedding space. Scenes differing by one vehicle receive partial credit rather than being treated as completely different.
    \item \textbf{Scalable supervision:} Soft labels are computed automatically from spatial structure without requiring human annotation, allowing to scale naturally to large datasets.
\end{itemize}


\subsection{Vehicle Segmentation}
\label{sec:segmentation}

Vehicle segmentation examines the spatial information captured by the radar encoder's patch-level features which capture where vehicles appear in the heatmap. We reshape the final layer-normalized patch tokens into a feature map $\mathbf{F}\in\mathbb{R}^{768\times14\times14}$ and apply a Progressive UPsampling (PUP) decoder~\cite{seg-pup}. Four convolutional stages with BN and ReLU, each followed by $2\times$ bilinear upsampling, progressively recover the $224\times224$ resolution. A final sigmoid produces the segmentation probability map $\hat{\mathbf{M}}$. During training, the frozen encoder's patch features are passed to the segmentation head, keeping the pretrained representations intact while training only the decoder.

\noindent\textbf{Ground-truth masks:}
We project annotated vehicle centers onto the radar BEV image grid and construct Gaussian-smoothed masks around these locations. In simulation, vehicle positions are obtained from CARLA's actor states; on RADIATE, they are derived from annotated bounding-box centers.

The head minimizes a weighted Dice and binary cross-entropy objective, averaged over images,
\begin{equation*}
    \mathcal{L}_{\mathrm{center}}=0.6\mathcal{L}_{\mathrm{Dice}}+ 0.4\mathcal{L}_{\mathrm{BCE}}
\end{equation*}
where
\begin{equation*}
 \mathcal{L}_{\mathrm{Dice}}=1-\frac{2\sum_p\widehat M_p M_p+\epsilon}{\sum_p\widehat M_p+\sum_p M_p+\epsilon}
\end{equation*}
Simulation probes train the decoder with frozen radar features whereas for RADIATE, we fine-tune the encoder and decoder together with different learning rates.

\subsection{Vehicle Bounding Box Detection}
\label{sec:detection-head}
To evaluate real-world instance localization, we attach a rotated Faster R-CNN detector~\cite{fasterrcnn2015} to the pretrained radar encoder. A channel projection, learned upsampling and pooling form a multi-scale feature pyramid. A rotated region-proposal network~\cite{rrpn2018} generates candidate boxes, whose RoI-aligned features support vehicle/background classification. We fine-tune on RADIATE dataset using binary cross-entropy for proposal objectness, cross-entropy for vehicle classification, and L1 regression losses on encoded box offsets at both the proposal and refinement stages.

\begin{table*}[!htbp]
\caption{Simulation captioning results. All metrics are computed per distance-sector cell and averaged within each distance bin.}
\label{tab:captioning-results}
    \centering
    \resizebox{\textwidth}{!}{
        \begin{tabular}{l|ccc|ccc|ccc|ccc}
        \hline
        \multirow{2}{*}{\textbf{Method}} & \multicolumn{3}{c|}{\textbf{0--10m}} & \multicolumn{3}{c|}{\textbf{10--20m}} & \multicolumn{3}{c|}{\textbf{20--30m}} & \multicolumn{3}{c}{\textbf{30--40m}} \\
        & \textbf{Prec} $\uparrow$ & \textbf{Rec} $\uparrow$ & \textbf{F1} $\uparrow$ 
        & \textbf{Prec} $\uparrow$ & \textbf{Rec} $\uparrow$ & \textbf{F1} $\uparrow$ 
        & \textbf{Prec} $\uparrow$ & \textbf{Rec} $\uparrow$ & \textbf{F1} $\uparrow$ 
        & \textbf{Prec} $\uparrow$ & \textbf{Rec} $\uparrow$ & \textbf{F1} $\uparrow$ \\ \hline
        Vanilla CLIP~\cite{clip} 
            & 0.961 & 0.926 & 0.943 
            & 0.900 & 0.817 & 0.844 
            & 0.742 & 0.669 & 0.693 
            & 0.673 & 0.529 & 0.577 \\ \hline
        \textbf{SG-CLIP ($\alpha=0.25$)} 
            & 0.983 & 0.981 & 0.982 
            & 0.929 & 0.916 & 0.922 
            & 0.849 & 0.816 & 0.829 
            & 0.870 & 0.844 & 0.856 \\
        \textbf{SG-CLIP ($\alpha=1.0$)}  
            & \textbf{0.986} & \textbf{0.985} & \textbf{0.985} 
            & \textbf{0.938} & \textbf{0.930} & \textbf{0.934} 
            & 0.863 & \textbf{0.858} & \textbf{0.861} 
            & 0.887 & \textbf{0.848} & \textbf{0.867} \\
        \textbf{SG-CLIP ($\alpha=4.0$)}  
            & 0.985 & 0.981 & 0.983 
            & 0.927 & 0.918 & 0.922 
            & \textbf{0.875} & 0.834 & 0.852 
            & \textbf{0.888} & 0.848 & 0.867 \\ \hline
        \end{tabular}
    }
\end{table*}

\subsection{Generative Scene Description}
\label{sec:captioning}

Generative captioning examines whether the global scene embedding encodes the structured vehicle distributions described during pretraining. Following ClipCap~\cite{clipcap}, a Transformer mapping network projects this radar embedding into prefix tokens for GPT-2~\cite{gpt2}. The prefix acts as a soft prompt, conditioning the decoder to describe vehicle counts and their spatial relations. Teacher forcing trains the captioning branch with
\begin{equation*}
    \mathcal{L}_{\mathrm{caption}}=-\sum_{t=1}^{L}\log p(c_t\mid\mathbf{p}_{1:K},c_{<t}),
\end{equation*}
where $\mathbf{p}_{1:K}$ is the learned radar-conditioned prefix. We evaluate this branch in simulation with the radar encoder frozen, probing the spatial information accessible from its global embedding. At inference, the radar embedding alone conditions autoregressive caption generation. Decoded vehicle counts are evaluated by distance and spatial relation, complementing the localization tasks supported by the patch features.

\section{Experiments}
\label{sec:experiments}

We evaluate RLM along three complementary axes: spatial information accessible through frozen-feature probes in simulation, transfer to real-world vehicle detection and segmentation, and performance under limited real-world annotation budgets. The real-world experiments on RADIATE~\cite{radiate} test the practical value of pretraining, while captioning and segmentation in simulation examine the global and local structure of the learned representations. Complete training configurations, dataset partitions, and implementation details are provided in the supplementary material.

\subsection{Experimental Setup}
\label{sec:real_protocol}

\paragraph{Datasets and settings:}
Simulation experiments use the dataset described in Section~\ref{sec:dataset}. Real-world transfer uses a sequence-disjoint RADIATE~\cite{radiate} partition containing 20,753 training, 3,630 validation, and 8,614 test frames, with car, van, truck, and bus grouped into one vehicle category. All perception experiments use a single radar frame; temporal methods such as SIRA and TempoRadar use additional observations and fall outside this comparison~\cite{sira,temporadar}.

\paragraph{Baselines:}
The representation comparisons use ViT-B/16 encoders initialized from OpenAI CLIP~\cite{clip}. \textit{OpenAI CLIP} adapts the released visual encoder directly to the downstream task, without additional simulation pretraining. \textit{Vanilla CLIP} first pretrains on our simulated radar-caption pairs using standard one-hot contrastive targets. \textit{RankCLIP}~\cite{rank-clip} uses these pairs with additional cross-modal and intra-modal ranking consistency. \textit{RADICaL}~\cite{bootstrap-radar} denotes our radar-only adaptation to the same ViT backbone: two independently augmented radar views are aligned without captions or camera supervision. These comparisons distinguish the benefit of simulation pretraining from the additional contribution of SG-CLIP's spatial targets.

For detection, the ViT encoders share the feature adapter, rotated detector, and head initialization within each downstream seed. Task-specific architecture baselines are BBAVectors~\cite{bbavectors2021} with ResNet-18 and rotated Faster R-CNN~\cite{fasterrcnn2015} with ResNet-50. For RADIATE segmentation, we include U-Net~\cite{unet} with ResNet-34.

\paragraph{Pretraining and adaptation:}
All radar-language pretraining variants use AdamW with an initial learning rate of $10^{-5}$ for 50 epochs. RLM uses a global contrastive batch of 640 on 4 NVIDIA A100 GPUs, requiring approximately two days. In our experiments, we found that larger batches improve contrastive learning by providing more diverse vehicle distributions and spatial relationships. Additionally, to accommodate the sensor-domain shift from the simulated TI radar configuration (Section~\ref{sec:dataset}) to RADIATE's Navtech scanning radar, we jointly fine-tune the pretrained encoder (low lr) and task head. The adaptation uses 30 epochs and encoder/head learning rates of $10^{-6}/10^{-3}$, respectively.

\paragraph{Limited-label training.}
Detection uses 5\%, 10\%, 20\%, 50\%, and 100\% of the labeled training frames. Within each seed, methods use the same nested subsets of contiguous blocks from training sequences, while validation and test sets remain complete. Training schedules for each annotation budget are detailed in the supplementary material


\begin{figure*}[!t]
    \centering
    \begin{subfigure}[b]{0.24\textwidth}
        \centering
        \includegraphics[width=\linewidth]{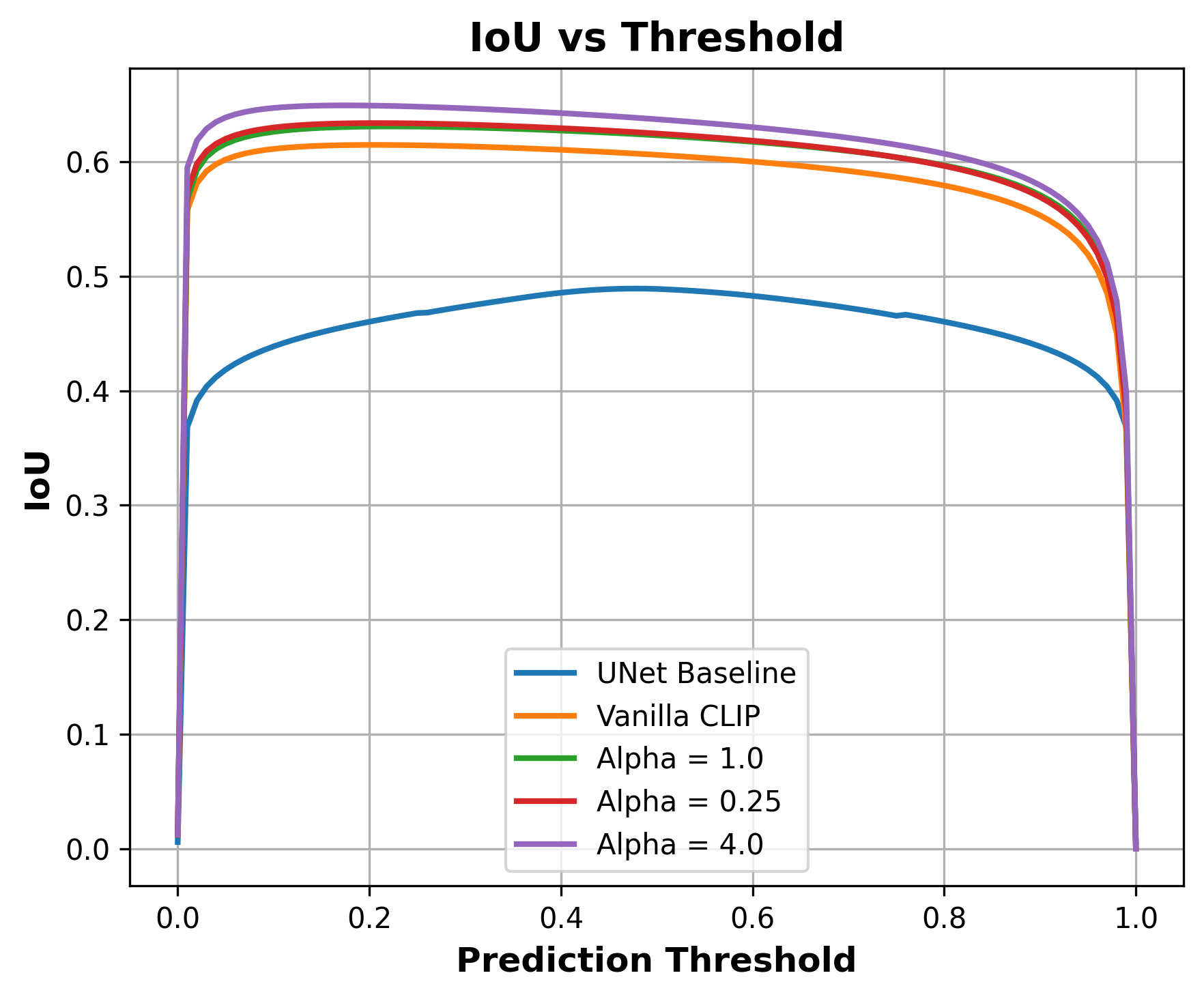}
        \caption{Simulation IoU}
        \label{fig:iou-sim}
    \end{subfigure}
    \hfill
    \begin{subfigure}[b]{0.24\textwidth}
        \centering
        \includegraphics[width=\linewidth]{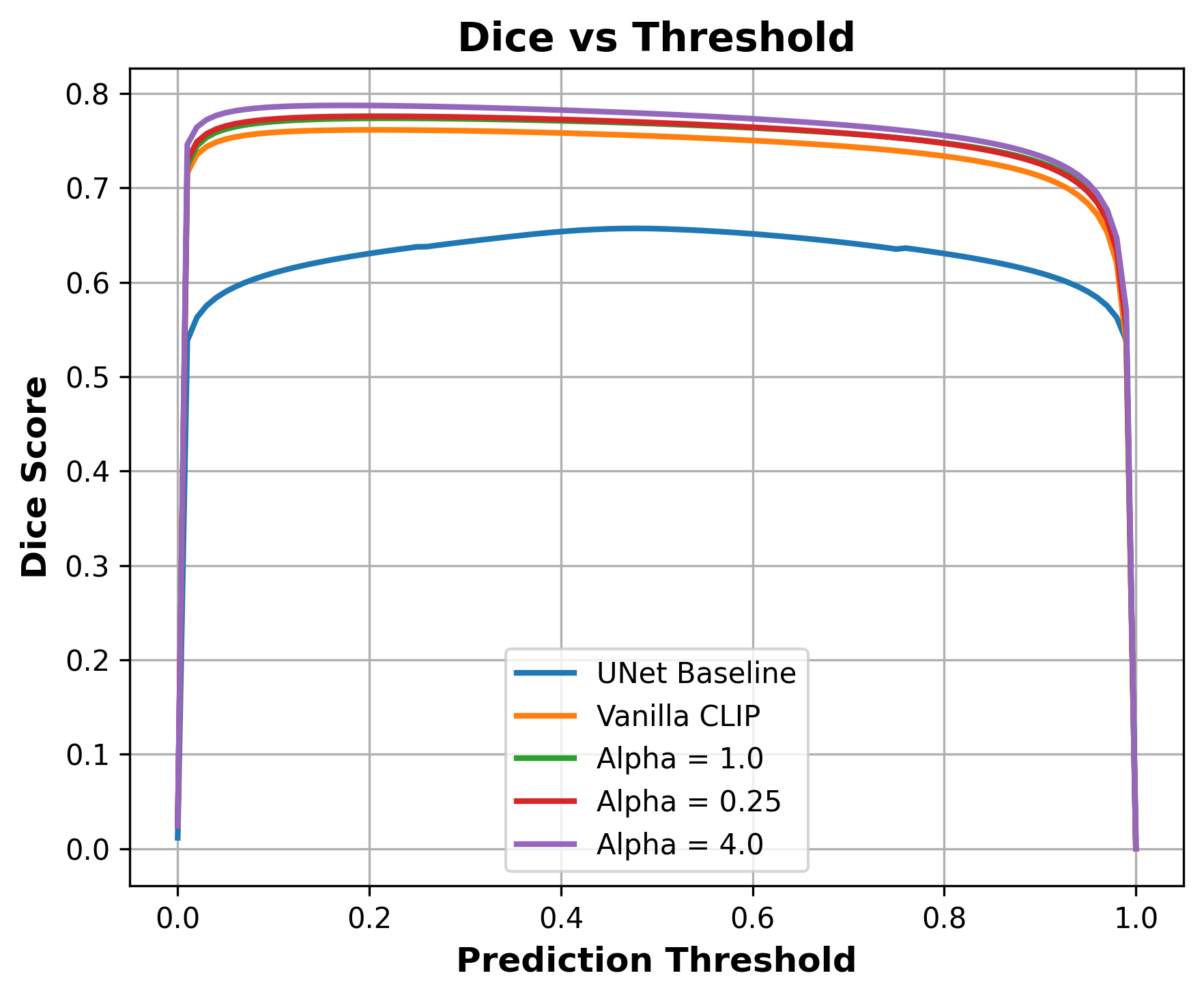}
        \caption{Simulation Dice}
        \label{fig:dice-sim}
    \end{subfigure}
    \hfill
    \begin{subfigure}[b]{0.24\textwidth}
        \centering
        \includegraphics[width=\linewidth]{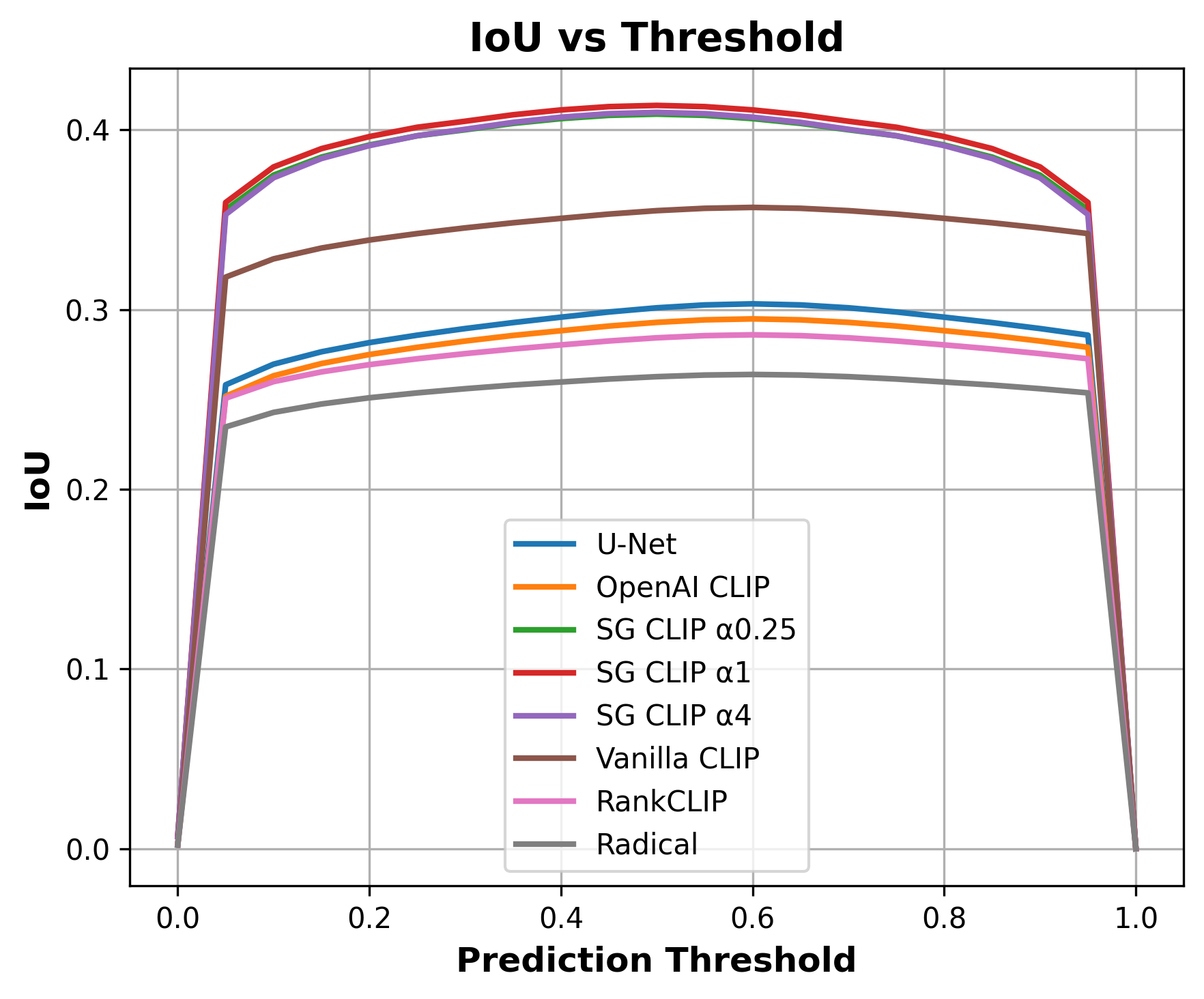}
        \caption{RADIATE IoU}
        \label{fig:iou-radiate}
    \end{subfigure}
    \hfill
    \begin{subfigure}[b]{0.24\textwidth}
        \centering
        \includegraphics[width=\linewidth]{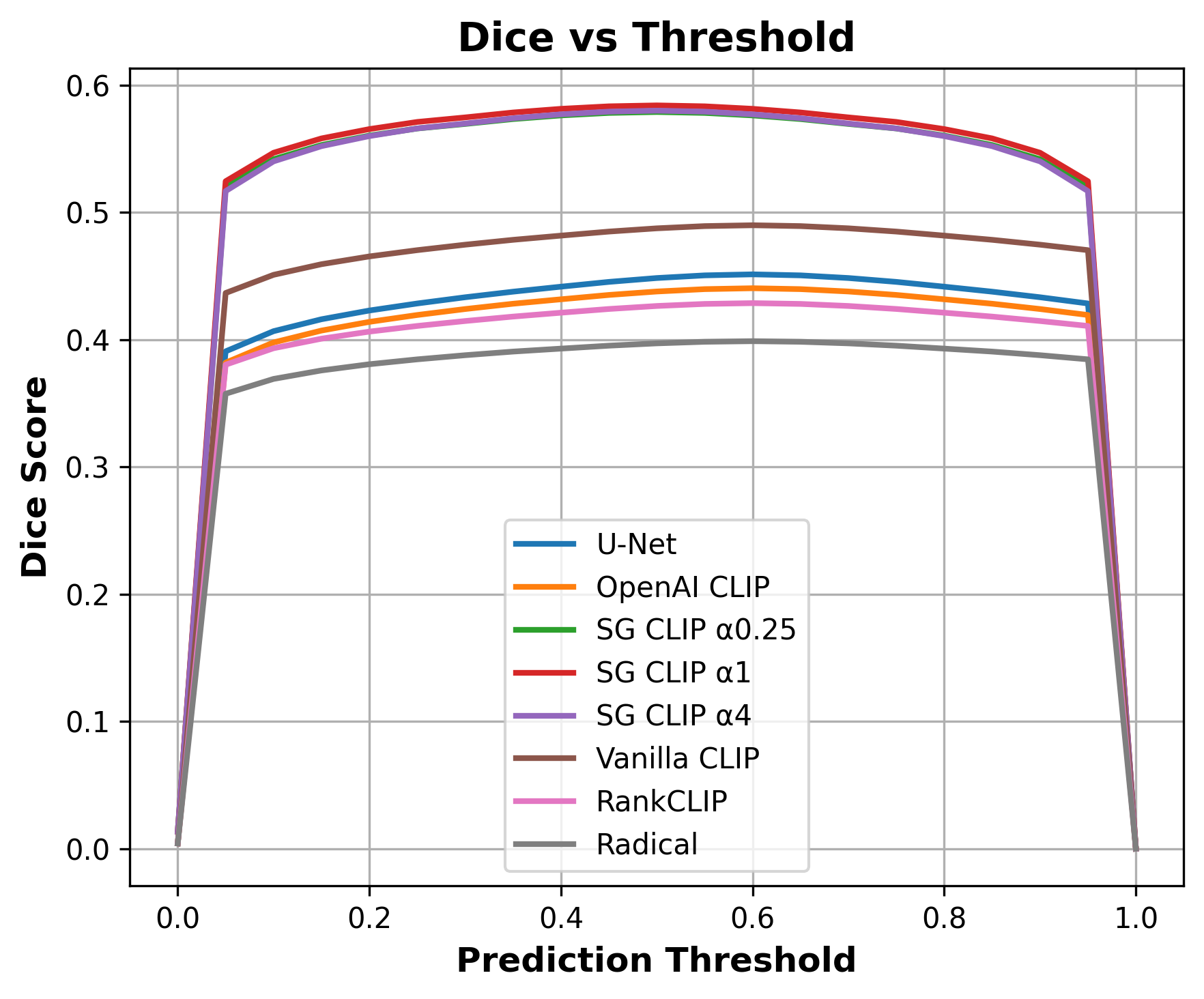}
        \caption{RADIATE Dice}
        \label{fig:dice-radiate}
    \end{subfigure}
    \caption{Vehicle segmentation IoU and Dice across prediction thresholds: (a,b) simulation; (c,d) RADIATE. ViT encoders are frozen for the simulation probes and jointly fine-tuned with very low learning rate on RADIATE. Each curve represents one evaluated checkpoint.}
    \label{fig:segmentation}
\end{figure*}

\paragraph{Evaluation metrics.}
\label{sec:metrics}
Captioning requires assigning the correct vehicle counts to spatial regions. Standard captioning metrics~\cite{bleu,meteor,cider} measure agreement with reference wording without explicitly evaluating these assignments. We therefore use localization-aware precision, recall and F1, extracting vehicle counts from generated descriptions with an LLM-assisted parser. For reference count $y$ and predicted count $\widehat y$ in each distance-sector cell, we compute
\begin{align}
    \mathrm{TP}&=\min(y,\widehat y),\nonumber\\
    \mathrm{FP}&=\max(0,\widehat y-y),\quad
    \mathrm{FN}=\max(0,y-\widehat y)\nonumber
    \label{eq:caption-count-metrics}
\end{align}
We aggregate these counts across test scenes within each cell to compute precision, recall and F1, then average the cell scores over spatial relations within each distance bin.
For bounding-box detection, we report AP50 as the primary metric and AP30 as a complementary measure, using oriented-box IoU thresholds of 0.50 and 0.30, respectively.
Vehicle segmentation is evaluated with IoU and Dice to measure mask overlap.

\subsection{Captioning in Simulation}
\label{sec:captioning_results}

With the radar encoder frozen, captioning tests whether its global representation retains vehicle counts and their spatial allocation. SG-CLIP improves count-based F1 over vanilla CLIP in every distance bin (Table~\ref{tab:captioning-results}), with the largest gains at longer ranges. At 30-40\,m, SG-CLIP with $\alpha=1$ achieves 0.867 F1 compared with 0.577 for vanilla CLIP, whereas the corresponding scores at 0-10\,m are 0.985 and 0.943. Both precision and recall improve in the farthest bin, indicating more accurate recovery of regional vehicle counts rather than simply generating more vehicle mentions. The distance-dependent gains show that spatially informed pretraining is particularly useful where the baseline's count accuracy degrades most.



\subsection{Vehicle Segmentation in Simulation}
\label{sec:segmentation_results}
Segmentation examines whether the benefit observed in the global captioning is also accessible from local patch features. With frozen encoders, SG-CLIP with $\alpha=4$ achieves 63.7\% IoU and 77.8\% Dice, compared with 60.6\% and 75.5\% for vanilla CLIP at threshold 0.5. The advantage persists across a broad threshold interval in Figure~\ref{fig:iou-sim} and Figure~\ref{fig:dice-sim}, showing that it is not specific to one operating point. These results indicate that scene-level spatial supervision also benefits local vehicle prediction.


\subsection{Real-World Vehicle Segmentation}
\label{sec:real_center}

Figures~\ref{fig:iou-radiate} and~\ref{fig:dice-radiate} show that SG-CLIP variants achieve higher measured scores than the compared baselines, indicating that the advantage is not confined to a single operating point. Vanilla CLIP is the strongest baseline, outperforming both OpenAI CLIP and the task-specific U-Net. This suggests that standard radar-language pretraining in simulation provides a useful initialization.

SG-CLIP achieves its best Dice at threshold 0.5, with $\alpha=1$ reaching 58.40\% Dice and IoU of 41.37\%. Comparing each model at its Dice-optimal threshold, this exceeds vanilla CLIP by 9.42 percentage points in Dice and 5.68 points in IoU. The three SG-CLIP variants differ by only 0.52 points in Dice and 0.49 points in IoU, suggesting that the measured gains persist across the tested spatial weightings rather than depending on a narrowly chosen $\alpha$.


\subsection{Real-World Vehicle Detection}
\label{sec:real_detection}

Table~\ref{tab:radiate-detection-all} separates the benefit of simulation pretraining from that of spatial supervision. With the same ViT detector, vanilla CLIP achieves 46.58\% mean AP50 compared with 42.67\% for OpenAI CLIP, showing that radar-language pretraining provides a useful initialization for real-world detection. SG-CLIP with $\alpha=1$ reaches 52.49\%, adding a further 5.91 percentage points over vanilla CLIP. The spatial objective therefore contributes beyond the benefit obtained from standard radar-text pretraining.

SG-CLIP also achieves higher mean AP50 than RankCLIP, RADICaL, and the task-specific architecture baselines under their reported training configurations. While$\alpha=1.0$ appears to have the highest score, $\alpha=0.25$ and $\alpha=4$ also do perform better than vanilla CLIP.

\begin{table}[!htbp]
    \caption{Full-label RADIATE detection. AP (\%), mean $\pm$ sample SD over three downstream seeds. ViT initializations share the detector; architecture baselines use their own models.}
    \label{tab:radiate-detection-all}
    \centering\small
    \setlength{\tabcolsep}{3pt}
    \begin{tabular}{@{}lcc@{}}
        \toprule
            Method & AP30 & AP50 \\
        \midrule
            \multicolumn{3}{l}{\emph{Shared ViT detector}} \\
            OpenAI CLIP & 62.36 $\pm$ 0.66 & 42.67 $\pm$ 1.03 \\
            Vanilla CLIP & 65.31 $\pm$ 2.15 & 46.58 $\pm$ 1.17 \\
            RankCLIP & 64.44 $\pm$ 0.70 & 45.80 $\pm$ 1.23 \\
            RADICaL (adapted) & 59.38 $\pm$ 1.49 & 40.88 $\pm$ 1.45 \\
            \textbf{SG-CLIP ($\alpha=0.25$)} & 66.43 $\pm$ 3.09 & 48.80 $\pm$ 1.65 \\
            \textbf{SG-CLIP ($\alpha=1$)} & \textbf{70.21 $\pm$ 1.43} & \textbf{52.49 $\pm$ 1.03} \\
            \textbf{SG-CLIP ($\alpha=4$)} & 67.02 $\pm$ 2.60 & 49.25 $\pm$ 0.86 \\
        \midrule
            \multicolumn{3}{l}{\emph{Task-specific architectures}} \\
            BBAVectors & 62.09 $\pm$ 1.18 & 42.63 $\pm$ 1.40 \\
            \shortstack[l]{Rotated Faster\\R-CNN} & 65.27 $\pm$ 0.58 & 41.74 $\pm$ 0.80 \\
        \bottomrule
    \end{tabular}
\end{table}

\begin{table*}[!ht]
    \caption{RADIATE vehicle detection with limited training labels. Columns indicate the fraction of labeled training frames; all budgets use the same full test split. Results are mean AP (\%) across downstream seeds.}
    \label{tab:radiate-label-budgets}
    \centering\small
    \setlength{\tabcolsep}{5pt}
    \begin{tabular*}{\textwidth}{@{\extracolsep{\fill}}lrrrrrrrr@{}}
        \toprule
            \multirow{2}{*}{Method} & \multicolumn{2}{c}{5\% labels} & \multicolumn{2}{c}{10\% labels} & \multicolumn{2}{c}{20\% labels} & \multicolumn{2}{c}{50\% labels} \\
            & AP30 & AP50 & AP30 & AP50 & AP30 & AP50 & AP30 & AP50 \\
        \midrule
            \multicolumn{3}{l}{\emph{Shared ViT detector}} \\
            OpenAI CLIP & 37.62 & 18.21 & 45.17 & 22.58 & 49.69 & 28.40 & 59.41 & 38.41 \\
            Vanilla CLIP & 41.95 & 20.52 & 47.58 & 24.77 & 53.61 & 31.59 & 60.65 & 39.78 \\
            RankCLIP & 40.84 & 19.74 & 48.28 & 25.55 & 53.95 & 31.48 & 60.18 & 39.25 \\
            RADICaL (adapted) & 29.21 & 16.27 & 33.74 & 20.54 & 39.37 & 25.34 & 49.56 & 31.72 \\
            SG-CLIP ($\alpha=1$) & \textbf{43.62} & \textbf{24.82} & \textbf{52.32} & \textbf{29.92} & \textbf{58.39} & \textbf{36.22} & \textbf{65.21} & \textbf{44.48} \\
        \midrule
            \multicolumn{3}{l}{\emph{Task-specific architectures}} \\
            BBAVectors & 34.38 & 18.96 & 43.79 & 24.95 & 49.61 & 31.38 & 57.54 & 39.48 \\
            Faster R-CNN & 40.03 & 20.54 & 46.15 & 26.26 & 51.66 & 29.99 & 55.32 & 36.31 \\
        \bottomrule
    \end{tabular*}
\end{table*}

\begin{table*}[!b]
    \caption{Vehicle detection with AP50 scores on RADIATE under adverse Weather; columns indicate the fraction of training labels used.}
    \label{tab:radiate-adverse-detection}
    \centering\small
    \setlength{\tabcolsep}{5pt}
    \begin{tabular*}{\textwidth}{@{\extracolsep{\fill}}lccccc@{}}
        \toprule
            \multirow{2}{*}{Method} & \multicolumn{5}{c}{Real training labels} \\
        \cmidrule(lr){2-6}
            & 5\% & 10\% & 20\% & 50\% & 100\% \\
        \midrule
            \multicolumn{3}{l}{\emph{Shared ViT detector}} \\
            OpenAI CLIP
            & 20.6 $\pm$ 3.7
            & 28.7 $\pm$ 1.0
            & 33.7 $\pm$ 2.5
            & 45.9 $\pm$ 3.4
            & 56.3 $\pm$ 2.4 \\
            Vanilla CLIP
            & 28.8 $\pm$ 3.0
            & 33.1 $\pm$ 2.9
            & 41.1 $\pm$ 4.3
            & 49.1 $\pm$ 2.1
            & 58.9 $\pm$ 2.2 \\
            RankCLIP
            & 26.8 $\pm$ 1.8
            & 33.7 $\pm$ 1.7
            & 34.5 $\pm$ 1.2
            & 41.4 $\pm$ 1.2
            & 46.7 $\pm$ 1.0 \\
            RADICaL (adapted)
            & 17.4 $\pm$ 2.8
            & 24.9 $\pm$ 1.3
            & 31.1 $\pm$ 6.0
            & 35.3 $\pm$ 1.0
            & 48.5 $\pm$ 1.1 \\
            \textbf{SG-CLIP ($\alpha=1$)}
            & \textbf{31.3 $\pm$ 2.9}
            & \textbf{38.6 $\pm$ 1.3}
            & \textbf{44.7 $\pm$ 1.5}
            & \textbf{51.7 $\pm$ 0.8}
            & \textbf{61.2 $\pm$ 1.0} \\
        \midrule
            \multicolumn{3}{l}{\emph{Task-specific architectures}} \\
            BBAVectors
            & 22.3 $\pm$ 1.8
            & 28.3 $\pm$ 1.4
            & 36.6 $\pm$ 1.3
            & 42.7 $\pm$ 1.6
            & 48.7 $\pm$ 1.6 \\
            Rotated Faster R-CNN
            & 25.3 $\pm$ 3.3
            & 34.4 $\pm$ 6.3
            & 35.4 $\pm$ 3.1
            & 40.4 $\pm$ 4.6
            & 54.4 $\pm$ 1.8 \\
        \bottomrule
    \end{tabular*}
\end{table*}

\subsection{Adaptation with Limited Real-World Labels}
\label{sec:label-efficiency}

SG-CLIP achieves the highest mean AP30 and AP50 at every partial-label budget in Table~\ref{tab:radiate-label-budgets}. At 10\% labels, its AP50 is 29.92\%, compared with 24.77\% for vanilla CLIP and 22.58\% for OpenAI CLIP. This comparison shows that both simulation pretraining and spatial supervision remain useful when real-world annotations are limited. However, the additional gain over vanilla CLIP varies with the budget: it is 5.15 percentage points at 10\% labels but only 4.30 points at 5\%. The benefit persists across the tested budgets without increasing monotonically as annotations become scarcer.



\subsection{Detection under Adverse Weather}
\label{sec:adverse_detection}

Table~\ref{tab:radiate-adverse-detection} tests whether the pretraining benefit extends to the recorded snow, fog, and rain sequences. With full supervision, vanilla CLIP achieves 58.9\% AP50 compared with 56.3\% for OpenAI CLIP, while SG-CLIP reaches 61.2\%. Much of the improvement is therefore already obtained through standard simulation radar-language pretraining, with spatial supervision providing an additional gain. The results support the usefulness of simulation pretraining on these challenging sequences.

\section{Conclusions and Future Directions}
\label{sec:conclusion}

We presented RLM, a framework for learning radar representations through structured spatial language supervision in simulation. SG-CLIP uses the vehicle distributions underlying those descriptions to define soft contrastive targets. The resulting radar-specific pretraining supports real-world oriented detection after supervised adaptation, including under limited annotation budgets, while dense center-map prediction and simulation scene generation provide complementary evaluations of the learned spatial features. The results establish a useful role for simulation-generated radar-language supervision in real perception and motivate its use as an initialization for task-specific learning.

Our experiments use vehicle distributions and transfer to one real radar dataset. Dense targets represent vehicle centers rather than full object silhouettes, and language generation is evaluated in simulation. The comparison with vanilla radar-text pretraining shows that useful transfer is not exclusive to SG-CLIP; identifying the causal contribution of language separately from structured labels and pretraining exposure remains an interesting direction. Future work can combine this initialization with temporal feature aggregation, extend evaluation across sensors and datasets~\cite{oxford-radar-1, oxford-radar-2, boreas, mulran}, and investigate real-radar descriptions and additional road users.


\raggedbottom
{\small
\bibliographystyle{ieeenat_fullname}
\bibliography{main}
}

\clearpage
\appendix
\addtocontents{toc}{\protect\setcounter{tocdepth}{2}}
\setcounter{table}{0}\setcounter{figure}{0}
\renewcommand{\thetable}{S\arabic{table}}
\renewcommand{\thefigure}{S\arabic{figure}}
\renewcommand{\theHtable}{S\arabic{table}}
\renewcommand{\theHfigure}{S\arabic{figure}}
\let\suppOrigSection\section
\renewcommand{\section}{\FloatBarrier\suppOrigSection}
\setcounter{topnumber}{4}
\setcounter{dbltopnumber}{4}
\renewcommand{\topfraction}{0.9}
\renewcommand{\dbltopfraction}{0.9}
\renewcommand{\textfraction}{0.08}
\makeatletter
\setlength{\@fptop}{0pt}
\setlength{\@fpsep}{12pt}
\setlength{\@dblfptop}{0pt}
\setlength{\@dblfpsep}{12pt}
\makeatother

\onecolumn
\begingroup
\nolinenumbers
\begin{center}
{\Large\textbf{\thetitle}\par}
\vspace{0.4em}
{\large Supplementary Material\par}
\end{center}
\vspace{0.5em}
\noindent
This supplement provides details of the simulated radar-language dataset, model architectures, training and evaluation protocols, and baseline implementations. It also includes per-cell captioning results, the simulation segmentation probe, and qualitative results in simulation and on RADIATE.
\par\vspace{0.5em}
\endgroup

\begingroup
\nolinenumbers
\setcounter{tocdepth}{2}
\hypersetup{linkcolor=black}
\fontsize{10}{11.5}\selectfont
\setlength{\parskip}{0pt}
\makeatletter
\renewcommand{\contentsname}{Supplementary Contents}
\renewcommand{\l@section}[2]{%
  \addvspace{3pt}%
  \begingroup\bfseries\@dottedtocline{1}{0pt}{1.8em}{#1}{#2}\endgroup}
\renewcommand{\l@subsection}[2]{\@dottedtocline{2}{1.8em}{2.8em}{#1}{#2}}
\renewcommand{\@dotsep}{2.5}
\makeatother
\tableofcontents
\endgroup
\clearpage

\twocolumn[{%
\begin{minipage}{\textwidth}
\centering
\captionsetup{hypcap=false}
\includegraphics[width=0.82\linewidth]{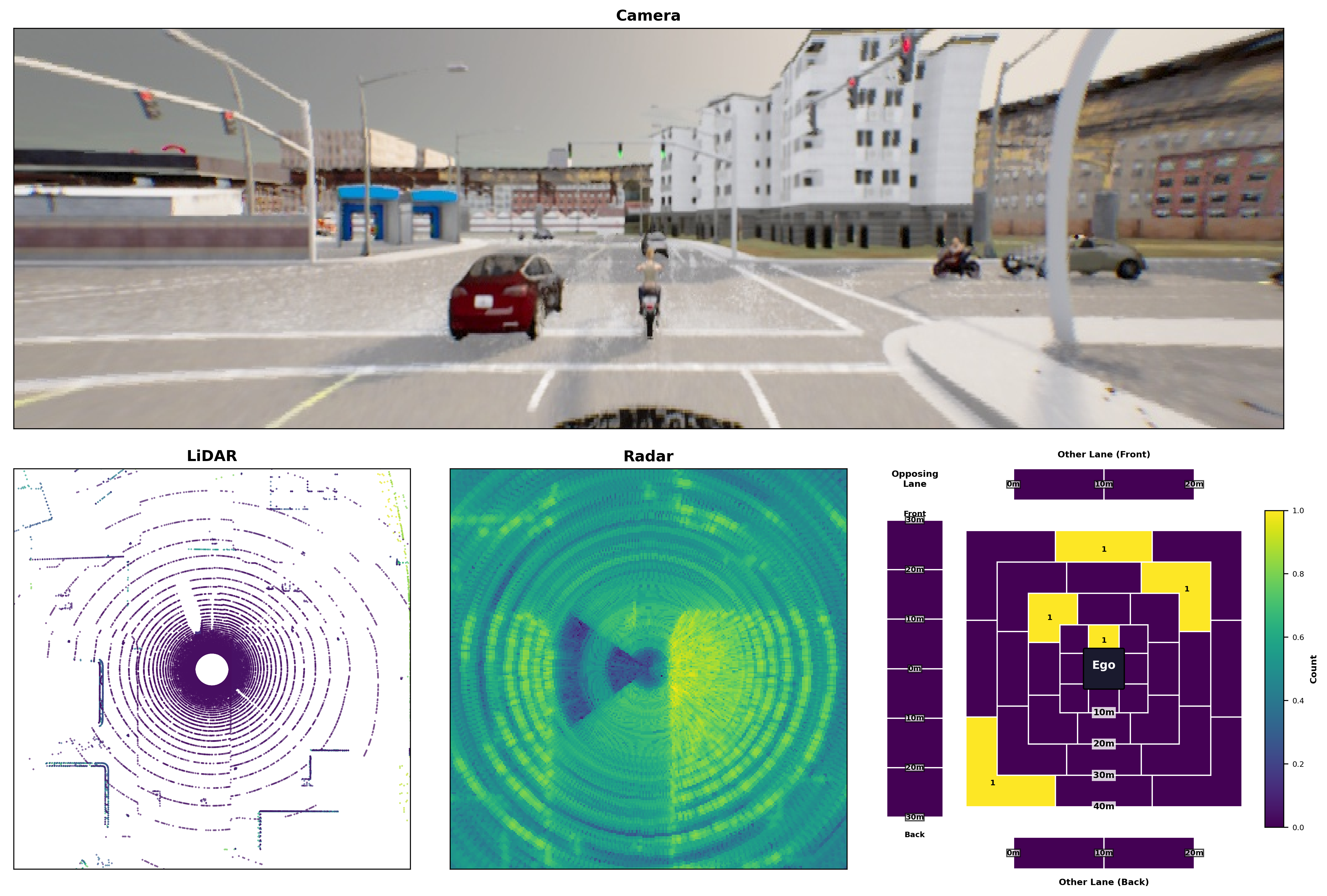}
\captionof{figure}{A representative data sample from our CARLA dataset. \textbf{Top:} camera reference view. \textbf{Bottom left to right:} LiDAR bird's-eye view, radar range-angle heatmap, and the ego-centric spatial grid annotation showing per-sector vehicle counts across four distance bins (0--40\,m). This scene depicts a Scenario~S3 (VRU crossing) event: a cyclist and motorcycle are visible in the in-lane front sector within 10\,m, with additional vehicles in the opposing lane at 20--30\,m.}
\label{fig:data_sample}
\end{minipage}
\vspace{0.8em}
}]

\section{CARLA Dataset}
\label{sec:supp-dataset}

Real-world annotated radar datasets remain scarce and small in scale, making large-scale supervised pre-training infeasible directly from real data. To overcome this, we leverage the C-Shenron radar simulator~\cite{c-shenron, c-shenron-demo}, an open-source, high-fidelity implementation that accurately models the propagation behaviour of automotive radars within the CARLA driving simulator~\cite{carla-sim}. Crucially, CARLA provides perfect ground-truth spatial annotations for every actor in the scene (as illustrated in Figure~\ref{fig:data_sample}), something that is prohibitively expensive to obtain at scale in the real world. Our dataset therefore, serves as a structured pre-training foundation: the model learns spatially-grounded radar representations in simulation, which are then transferred and validated on real-world radar data. The correspondence between C-Shenron and real radar measurements is studied in~\cite{c-shenron, c-shenron-demo}; here, the RADIATE experiments test whether representations learned in simulation transfer to real radar data.

To expose pre-training to varied spatial configurations, we collect data across a range of driving scenarios following the scenario distribution of~\cite{neat}, including adversarial traffic conflicts, unsignalized junctions, vulnerable road user crossings, adverse weather, and lighting conditions from noon to night. The radar sensor is configured to simulate a 77\,GHz Texas Instruments cascaded automotive radar system with 86 virtual antennas, providing realistic Range-Azimuth-Doppler measurements. In total, we collected 879,383 radar-caption frame pairs across distinct CARLA maps and several weather presets, described in detail in the following subsections.

\begin{figure}[t]
    \centering
    \includegraphics[width=\columnwidth]{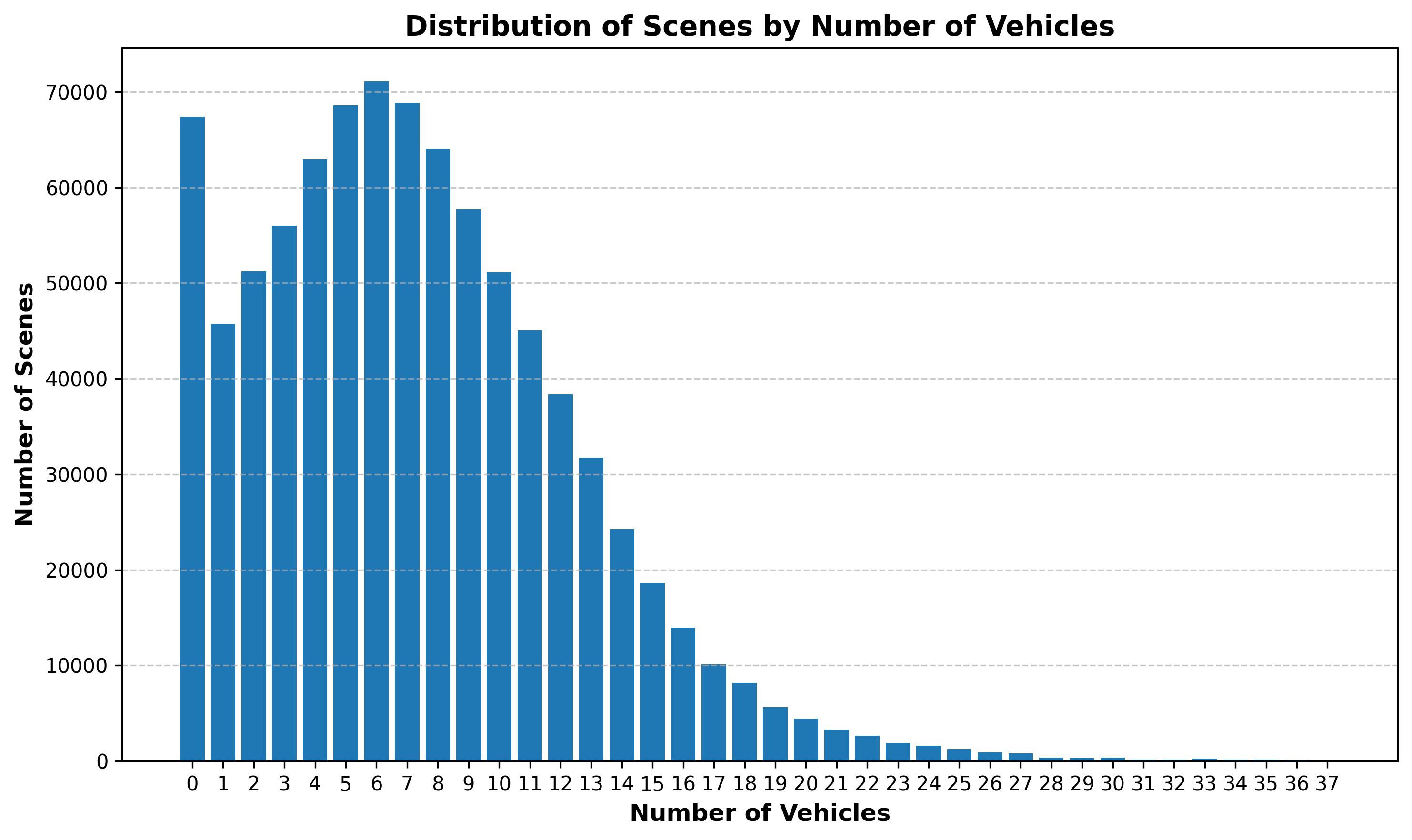}
    \caption{Number of vehicles per scene in the simulated dataset. Each scene corresponds to a radar frame.}
    \label{fig:num-vehicle-dist}
\end{figure}

Figure~\ref{fig:num-vehicle-dist} shows the distribution of frames by the number of vehicles in the scene.

\subsection{Simulation Environment}

We collect data across all eight CARLA maps available in CARLA 0.9.10, spanning residential, urban, highway, and rural environments to ensure broad coverage of driving conditions. Table~\ref{tab:towns} summarises the road network statistics for each map.

\begin{table}[!h]
    \centering
    \caption{CARLA map properties used for data collection.}
    \label{tab:towns}
    \resizebox{\linewidth}{!}{%
        \begin{tabular}{llcccc}
            \toprule
            \textbf{Map} & \textbf{Environment} & \textbf{Roads} & \textbf{Junctions} & \textbf{Max Lanes} & \textbf{Length} \\
            \midrule
            Town01   & Small-scale urban residential                        & 122 & 12 & 3  & 4.2\,km  \\
            Town02   & Very small urban residential                         & 84  & 8  & 3  & 2.0\,km  \\
            Town03   & Mixed urban (roundabout, elevated section)           & 369 & 35 & 10 & 11.3\,km \\
            Town04   & Mixed (urban core + highway loop)                    & 276 & 27 & 7  & 10.8\,km \\
            Town05   & Urban grid with elevated section                     & 323 & 21 & 7  & 9.2\,km  \\
            Town06   & Highway / motorway (Michigan left intersections)     & 200 & 23 & 10 & 8.5\,km  \\
            Town07   & Rural (narrow winding roads)                         & 277 & 31 & 9  & 5.4\,km  \\
            Town10HD & Dense urban, high-detail city                        & 108 & 9  & 8  & 3.4\,km  \\
            \bottomrule
        \end{tabular}
    }
\end{table}

To maximise scenario diversity and cover corner cases that a radar sensor may encounter, data collection is structured around adversarial scenario types drawn from the CARLA Leaderboard benchmark~\cite{neat}. Each scenario injects a distinct traffic conflict at a trigger waypoint, ensuring the model is exposed to varied spatial configurations of actors beyond routine lane-following. Table~\ref{tab:scenarios_adversarial} summarises each adversarial scenario, its NHTSA pre-crash typology, a brief description of the injected event, and the maps on which it is collected.

\begin{table}[!h]
    \centering
    \caption{Adversarial scenario types used for data collection. Town abbreviations: T01-T07 = Town01-Town07, T10 = Town10HD.}
    \label{tab:scenarios_adversarial}
    \resizebox{0.8\linewidth}{!}{%
        \renewcommand{\arraystretch}{1.2}
        \begin{tabular}{lp{2.2cm}p{6.0cm}p{2.2cm}}
            \toprule
            \textbf{ID} 
                & \makecell[l]{\textbf{NHTSA}\\\textbf{Typology}}
                & \textbf{Description} 
                & \textbf{Maps} \\
            \midrule
            S1  & \makecell[l]{Control Loss /\\Road Surface\\Hazard}
                & Triggered on any straight road section; debris and steering noise injected.
                & T01--T07, T10 \\[4pt]
            \midrule
            S3  & \makecell[l]{VRU Crossing\\Without Prior\\Warning}
                & Triggered when ego time-to-arrival to the VRU reaches 10\,s.
                & T01--T07, T10 \\[4pt]
            \midrule
            S4  & \makecell[l]{Turning at\\Intersection:\\Cyclist\\Conflict}
                & Triggered when ego comes within ${\sim}13$\,m of the junction exit on a left or right turn.
                & T01--T07, T10 \\[4pt]
            \midrule
            S7  & \makecell[l]{Signalised\\Intersection:\\Red Light\\Violation}
                & Triggered at signalised junction; conflicting lane receives simultaneous green. Subtypes: S7left, S7right, S7opposite.
                & T01--T07, T10 \\[4pt]
            \midrule
            S8  & \makecell[l]{Left Turn: \\Oncoming\\Encroachment}
                & Triggered only when next route maneuver is a left turn (S8left).
                & T01--T07, T10 \\[4pt]
            \midrule
            S9  & \makecell[l]{Right Turn:\\Cross-Traffic\\Conflict}
                & Triggered only when next route maneuver is a right turn (S9right).
                & T01--T07, T10 \\[4pt]
            \midrule
            S10 & \makecell[l]{Unsignalised\\Intersection}
                & Triggered at any unsignalised junction; no adversary spawned.
                & T03--T07, T10 \\
            \bottomrule
        \end{tabular}
    }
\end{table}

In addition to the adversarial scenarios above, we include four route-only configurations (Table~\ref{tab:scenarios_routes}) that collect pure lane-following data under dense ambient traffic with no injected events. These routes traverse multi-lane highway geometry and are named by the directional maneuver sequence encoded in their waypoint yaw progressions.

\begin{table}[tp]
    \centering\small
    \setlength{\tabcolsep}{2.5pt}
    \caption{Route-only lane-following configurations. No adversarial events 
             are injected; ego drives through ambient background traffic only.}
    \label{tab:scenarios_routes}
        \begin{tabular}{@{}l>{\raggedright\arraybackslash}p{1.7cm}>{\raggedright\arraybackslash}p{3.4cm}>{\raggedright\arraybackslash}p{1.6cm}@{}}
            \toprule
            \textbf{ID} 
                & \thead{\textbf{Maneuver}\\\textbf{Sequence}}
                & \textbf{Description} 
                & \textbf{Maps} \\
            \midrule
            ll
                & Left $\rightarrow$ Left
                & Highway curves with a left-then-left waypoint sequence.
                & T04, T05, T06 \\[4pt]
            \midrule
            lr
                & Left $\rightarrow$ Right
                & Highway curves with a left-then-right waypoint sequence.
                & T03--T06, T10 \\[4pt]
            \midrule
            rl
                & Right $\rightarrow$ Left
                & Adjacent-lane mirror of lr; laterally offset by 
                  ${\sim}3.5$\,m.
                & T03--T06, T10 \\[4pt]
            \midrule
            rr
                & Right $\rightarrow$ Right
                & Adjacent-lane mirror of ll; start points offset by 
                  ${\sim}6$\,m.
                & T04, T05, T06 \\

            \bottomrule
        \end{tabular}
\end{table}

\textbf{Weather diversity.} To vary the simulated collection conditions, each data collection episode samples its weather configuration independently at runtime. Seven base presets covering the full precipitation spectrum are combined with six sun altitude levels (night at $-80^\circ$, twilight at $0^\circ$, dawn at $5^\circ$, sunset at $15^\circ$, morning at $35^\circ$, noon at $75^\circ$) and eight evenly-spaced sun azimuth angles ($0^\circ, 45^\circ, \ldots, 315^\circ$), yielding 336 nominal weather configurations. A Gaussian perturbation ($\sigma = 10^\circ$) is additionally applied to the sun altitude, making the sampling effectively continuous. Table~\ref{tab:weather} summarises the base presets; altitude and azimuth overrides are applied on top of each preset at episode initialisation.

\begin{table}[!h]
    \centering\small
    \setlength{\tabcolsep}{3pt}
    \caption{Base weather presets used during data collection. Sun altitude and azimuth are overridden per episode by the runtime sampler (336 nominal combinations, continuously perturbed).}
    \label{tab:weather}
        \begin{tabular}{@{}lccc@{}}
            \toprule
            \textbf{Preset} & \textbf{Cloudiness} & \textbf{Precipitation} & \makecell{\textbf{Base}\\\textbf{Altitude ($^\circ$)}} \\
            \midrule
            ClearNoon       & 15 & 0  & 75 \\
            CloudySunset    & 80 & 0  & 15 \\
            WetSunset       & 20 & 0  & 15 \\
            MidRainSunset   & 80 & 30 & 15 \\
            WetCloudySunset & 90 & 0  & 15 \\
            HardRainNoon    & 90 & 60 & 75 \\
            SoftRainSunset  & 90 & 15 & 15 \\
            \bottomrule
        \end{tabular}
\end{table}

\subsection{Actor State Extraction and Filtering}
For each simulation frame, we extract the complete state of all actors present in the CARLA world. Actors include vehicles, pedestrians, and traffic infrastructure elements such as traffic lights and signs. For each actor, we record: unique identifier, 3D location coordinates, transform (pitch, yaw, roll), velocity vector, acceleration vector, traffic light association status, current and ground-truth traffic light states, traffic sign type, and applicable speed limits. This comprehensive state representation enables us to filter and classify actors based on the ego vehicle's location in real time.

To focus on perception-relevant objects, we apply distance-based filtering relative to the ego vehicle's position. Actors located beyond 40 meters from the ego vehicle are excluded, matching the 40\,m extent of the caption grid. For lane-wise classification, we transform actor positions into the ego vehicle's local coordinate frame where the ego's forward direction defines the positive x-axis and the left direction defines the positive y-axis. Each vehicle is classified based on its motion direction relative to the ego vehicle using dot products of forward vectors: vehicles are labeled as "same direction" (dot product $> 0.5$), "opposing direction" (dot product $< -0.5$), or "other" (perpendicular or crossing traffic). Within the "same direction" category, lateral offsets determine lane assignment (in-lane, left-lane, or right-lane) using a configurable lane width threshold of 3 meters. Additionally, each vehicle is classified as being ahead, behind, or directly beside the ego vehicle based on the dot product between the ego's forward vector and the relative position vector to that vehicle. This ego-centric spatial encoding provides the foundation for generating structured, spatially-grounded scene descriptions.

\subsection{Natural Language Caption Generation}
The natural-language caption-generation pipeline uses GPT-4.1 mini to convert structured JSON representations into varied descriptions. The prompts constrain semantic content while allowing lexical variation. The prompting structure consists of two components: a system prompt that instructs the model to adopt the perspective of a driver describing the scene to a passenger, and a user prompt that provides the JSON data along with detailed key-explanation mappings.

\subsubsection{System Prompt}

The system prompt establishes the generation style and constraints:

\begin{quote}
\textit{"You are an expert at describing traffic scenes from the perspective of a driver talking to a passenger. Only output the final caption. Do not include explanations. Also do not use short word forms with apostrophe, use the full expansion of words only."}
\end{quote}

This prompt requests naturalistic descriptions without meta-commentary or contractions.

\subsubsection{Key-Explanation Mappings}

The user prompt includes comprehensive mappings that provide semantic context for each spatial encoding element. These mappings are organized into two categories: distance bins and angular sectors.

\textbf{Distance Bin Mappings:}
\begin{itemize}
    \item \texttt{0-10m}: "Objects within 0 to 10 meters from the ego vehicle. These are very close and may require immediate attention."
    \item \texttt{10-20m}: "Objects within 10 to 20 meters from the ego vehicle. These are close and should be monitored closely."
    \item \texttt{20-30m}: "Objects within 20 to 30 meters from the ego vehicle. These are at a moderate distance and should be noted."
    \item \texttt{30-40m}: "Objects within 30 to 40 meters from the ego vehicle. These are at a safe distance but still relevant."
\end{itemize}

\textbf{Angular Sector and Feature Mappings:}
\begin{itemize}
    \item \texttt{total\_vehicles}: "Total number of vehicles in the distance bin."
    \item \texttt{left\_lane\_front\_side}: "Vehicles in the left adjacent lane ahead of the ego vehicle."
    \item \texttt{left\_side}: "Vehicles directly to the left side of the ego vehicle."
    \item \texttt{left\_lane\_back\_side}: "Vehicles in the left adjacent lane behind the ego vehicle."
    \item \texttt{in\_lane\_front\_side}: "Vehicles directly ahead of the ego vehicle in the same lane."
    \item \texttt{in\_lane\_back\_side}: "Vehicles directly behind the ego vehicle in the same lane."
    \item \texttt{right\_lane\_front\_side}: "Vehicles in the right adjacent lane ahead of the ego vehicle."
    \item \texttt{right\_side}: "Vehicles directly to the right side of the ego vehicle."
    \item \texttt{right\_lane\_back\_side}: "Vehicles in the right adjacent lane behind the ego vehicle."
    \item \texttt{opposing\_lane\_front}: "Vehicles in the opposing lane ahead of the ego vehicle."
    \item \texttt{opposing\_lane\_back}: "Vehicles in the opposing lane behind the ego vehicle."
    \item \texttt{other\_lane\_front}: "Vehicles in perpendicular or intersecting lanes ahead of the ego vehicle."
    \item \texttt{other\_lane\_back}: "Vehicles in perpendicular or intersecting lanes behind the ego vehicle."
\end{itemize}

\subsubsection{Complete Prompt Example}

To illustrate the complete prompting structure, consider a scene where the ego vehicle is on a highway with vehicles in multiple lanes. The structured JSON data is first preprocessed: numerical values are converted to word form (e.g., 3 becomes "three"), and keys within each distance bin are randomly shuffled to prevent the model from learning positional biases. The complete user prompt combines the JSON data with the key explanations:

\begin{quote}
\textit{Here is the JSON data:}
\begin{verbatim}
{
  "0-10m": {
    "total_vehicles": "three",
    "in_lane_front_side": "one",
    "right_lane_back_side": "two"
  },
  "10-20m": {
    "total_vehicles": "five",
    "right_lane_front_side": "three",
    "in_lane_back_side": "one",
    "right_lane_back_side": "one"
  },
  "20-30m": {
    "total_vehicles": "four",
    "opposing_lane_front": "four"
  },
  "30-40m": {
    "total_vehicles": "two",
    "in_lane_front_side": "one",
    "right_lane_front_side": "one"
  }
}
\end{verbatim}

\textit{Here is the description of every key in the JSON data:}

\textit{Distance Bin - 0-10m: Objects within 0 to 10 meters from the ego vehicle. These are very close and may require immediate attention.}

\textit{[... additional distance bin descriptions ...]}

\textit{- total\_vehicles: Total number of vehicles in the distance bin.}

\textit{- in\_lane\_front\_side: Vehicles directly ahead of the ego vehicle in the same lane.}

\textit{- right\_lane\_back\_side: Vehicles in the right adjacent lane behind the ego vehicle.}

\textit{[... additional key descriptions ...]}

\textit{Given this data, generate natural language sentences describing the scene. Only provide the description. Do not explain how you arrived at it.}
\end{quote}

Additionally, we randomly vary whether to include the instruction "Do not add any distance bin numbers" to create captions with and without explicit distance references, further increasing linguistic diversity.

\subsubsection{Caption Generation Parameters}
To enhance caption diversity, we generate multiple caption variants for each radar frame using the following sampling parameters: temperature $T=0.5$ to balance creativity with consistency, top-p nucleus sampling with $p=0.9$ to constrain the probability mass while allowing varied word choices, maximum token length of 400 to accommodate detailed scene descriptions, and number of samples $n=5$ to provide multiple linguistic realizations of the same spatial structure. Section~\ref{sec:supp-current-pretraining} describes how template and generated captions are sampled during training.

\section{Method Details}
\label{sec:supp-architecture}

A single radar encoder is shared by all tasks. Its global representation is used for radar-language alignment and captioning, and its patch features are used for segmentation and bounding-box detection.

\subsection{Radar Encoder}
\label{sec:supp-radar-architecture}

The radar encoder is initialized from the released CLIP ViT-B/16 image encoder~\cite{clip,vit}. A $224\times224$ radar input produces one CLS token and 196 patch tokens on a $14\times14$ grid. The projected CLS token is used for the contrastive objective and as the captioning prefix input. For spatial heads, the CLS token is discarded and the final layer-normalized (\texttt{ln\_post}) patch tokens are reshaped into a $768\times14\times14$ feature map.

\subsection{Text Encoder}
\label{sec:supp-text-encoder}

Our structured captions enumerate vehicle counts across four distance bins and twelve relations and often exceed the 77-token context of the original CLIP text encoder. We therefore instantiate a CLIP model with a context length of 400 tokens, copy the pretrained ViT-B/16 visual weights, and train the text encoder from scratch with a newly initialized positional embedding (Table~\ref{tab:text_encoder}). The output at the end-of-text token passes through a final LayerNorm and a linear projection into the 512-dimensional shared embedding space.

\begin{table}[!h]
    \centering\small
    \caption{Text encoder. All choices other than the context length and initialization follow the CLIP ViT-B/16 text tower.}
    \label{tab:text_encoder}
    \begin{tabular}{@{}ll@{}}
        \toprule
        Parameter & Value \\
        \midrule
        Architecture & Causal Transformer \\
        Context length & 400 tokens \\
        Positional encoding & Learned, $400\times512$, $\mathcal{N}(0,0.01^2)$ \\
        Vocabulary size & 49,408 \\
        Width / layers / heads & 512 / 12 / 8 \\
        MLP inner dim. & 2,048, QuickGELU \\
        Output token & End-of-text position \\
        Projection & Linear $512\to512$, no bias \\
        Initialization & From scratch \\
        \bottomrule
    \end{tabular}
\end{table}

\subsection{SG-CLIP Objective}
\label{sec:supp-sgclip}

Algorithm~\ref{alg:sgclip} summarizes one SG-CLIP training step in the notation of the main paper. The soft targets are computed on the batch gathered across GPUs, so every frame is contrasted with all $N=640$ captions in the global batch. Because the targets depend only on the vehicle-count vectors, template and LLM-generated captions of the same scene receive identical targets. The paired caption always receives the largest weight in its row, since $d_{ii}=0$, and captions of scenes with identical count vectors receive the same weight as the paired caption. Vanilla CLIP is the special case $T=I$, in which only the paired caption is a positive.

\begin{algorithm}[t]
\caption{One SG-CLIP training step.}
\label{alg:sgclip}
\small
\textbf{Input:} radar frames $\{x_i\}_{i=1}^{N}$ with vehicle-count vectors $\{\mathbf{v}_i\}$; radar and text encoders $f_r$, $f_t$; kernel parameter $\alpha$; temperature~$\tau$.
\begin{enumerate}[leftmargin=1.6em,nosep,label=\arabic*:]
    \item $c_i \leftarrow$ template caption of $\mathbf{v}_i$ with probability $0.25$, otherwise one of its LLM-generated captions
    \item $\mathbf{z}_r^i \leftarrow f_r(x_i)$, \ $\mathbf{z}_t^i \leftarrow f_t(c_i)$ \hfill $\triangleright$ 512-d projections
    \item Gather $\{\mathbf{z}_r^i, \mathbf{z}_t^i, \mathbf{v}_i\}$ across GPUs
    \item $d_{ij} \leftarrow \sum_b\sum_s \bigl|v_i^{(b,s)}-v_j^{(b,s)}\bigr|$
    \item $T_{ij} \leftarrow \exp(-\alpha d_{ij}^2) \,/\, \sum_k \exp(-\alpha d_{ik}^2)$
    \item $S_{ij} \leftarrow \mathbf{z}_r^i\cdot\mathbf{z}_t^j \,/\, \bigl(\|\mathbf{z}_r^i\|\,\|\mathbf{z}_t^j\|\bigr)$
    \item $\mathcal{L}_{r\rightarrow t} \leftarrow -\frac{1}{N}\sum_i\sum_j T_{ij}\log\frac{\exp(S_{ij}/\tau)}{\sum_k\exp(S_{ik}/\tau)}$
    \item $\mathcal{L}_{t\rightarrow r} \leftarrow -\frac{1}{N}\sum_j\sum_i T_{ji}\log\frac{\exp(S_{ij}/\tau)}{\sum_k\exp(S_{kj}/\tau)}$
    \item Update $f_r$ and $f_t$ with AdamW on $\tfrac{1}{2}\bigl(\mathcal{L}_{r\rightarrow t}+\mathcal{L}_{t\rightarrow r}\bigr)$
\end{enumerate}
\end{algorithm}

\subsection{Captioning Mapper and Decoder}
\label{sec:supp-caption-architecture}

Following ClipCap~\cite{clipcap}, a Transformer mapping network projects the global radar embedding into a sequence of prefix tokens for GPT-2~\cite{gpt2}. The prefix acts as a soft prompt that conditions the autoregressive decoder on the radar scene. At inference, generation is conditioned on the radar embedding alone. The GPT-2 decoder is separate from the contrastive text encoder above.

\subsection{Segmentation Decoder}
\label{sec:supp-pup}

The Progressive UPsampling (PUP) decoder~\cite{seg-pup} maps the $768\times14\times14$ patch features to a $224\times224$ vehicle segmentation map through four stages of $3\times3$ convolution and $2\times$ bilinear upsampling (Table~\ref{tab:pup_decoder}). The total upsampling factor of $16$ inverts the ViT-B/16 patch size.

\begin{table}[!h]
    \centering\footnotesize
    \setlength{\tabcolsep}{3pt}
    \caption{PUP decoder. Each $3\times3$ convolution is followed by batch normalization and ReLU; upsampling is bilinear. The final $1\times1$ convolution is followed by a sigmoid.}
    \label{tab:pup_decoder}
    \begin{tabular}{@{}clcc@{}}
        \toprule
        \# & Operation & Channels & Output grid \\
        \midrule
        1 & Conv $3\times3$ & $768\to256$ & $14\times14$ \\
        2 & Upsample $2\times$ & $256$ & $28\times28$ \\
        3 & Conv $3\times3$ & $256\to256$ & $28\times28$ \\
        4 & Upsample $2\times$ & $256$ & $56\times56$ \\
        5 & Conv $3\times3$ & $256\to128$ & $56\times56$ \\
        6 & Upsample $2\times$ & $128$ & $112\times112$ \\
        7 & Conv $3\times3$ & $128\to64$ & $112\times112$ \\
        8 & Upsample $2\times$ & $64$ & $224\times224$ \\
        9 & Conv $1\times1$ & $64\to1$ & $224\times224$ \\
        \bottomrule
    \end{tabular}
\end{table}

\subsection{Bounding-Box Detector}
\label{sec:supp-box-head}

All ViT initializations (OpenAI CLIP, vanilla CLIP, RankCLIP, RADICaL and SG-CLIP) use the same detector. The $768\times14\times14$ patch features are projected to 256 channels by a $1\times1$ convolution with group normalization. Two learned $2\times$ upsampling stages and pooled lower-resolution features form a multi-scale pyramid, which feeds a rotated region-proposal network and RoI head~\cite{fasterrcnn2015,rrpn2018}. The detector predicts a single vehicle class with oriented boxes. Within each downstream seed, the adapter and detector heads share the same initialization across all encoders, so the encoder is the only difference between these runs.

\section{Training and Evaluation Details}
\label{sec:supp-current-protocols}

All perception models operate on a single radar frame, without camera images or additional radar frames. Table~\ref{tab:supp-settings-overview} collects the settings of the SG-CLIP pipeline; the subsections below give the details and the settings of the compared methods.

\begin{table}[!h]
\centering\footnotesize
\setlength{\tabcolsep}{3pt}
\caption{Settings of the SG-CLIP pipeline at a glance. Pretraining settings of the compared methods are listed in Table~\ref{tab:supp-current-pretrain-settings}.}
\label{tab:supp-settings-overview}
\begin{tabular}{@{}lc@{}}
\toprule
Setting & Value \\
\midrule
\multicolumn{2}{@{}l}{\textit{Data}} \\
CARLA train / val / test frames & 703,490 / 87,922 / 87,971 \\
RADIATE train / val / test frames & 20,753 / 3,630 / 8,614 \\
RADIATE crop & $75\times75$\,m \\
Radar input size & $224\times224$ \\
Distance bins / relations & 4 / 12 \\
\midrule
\multicolumn{2}{@{}l}{\textit{Architecture}} \\
Radar encoder & ViT-B/16 \\
Patch feature map & $768\times14\times14$ \\
Shared embedding dim. & 512 \\
Text context length & 400 tokens \\
\midrule
\multicolumn{2}{@{}l}{\textit{Pretraining}} \\
Kernel parameter $\alpha$ & 0.25 / 1 / 4 \\
Temperature $\tau$ & 0.07 \\
Global batch $N$ & 640 (4$\times$A100) \\
Template caption probability & 0.25 \\
Optimizer / LR / weight decay & AdamW / $10^{-5}$ / 0.05 \\
Epochs & 50 \\
\midrule
\multicolumn{2}{@{}l}{\textit{Segmentation}} \\
Loss $\mathcal{L}_{\mathrm{center}}$ & $0.6\mathcal{L}_{\mathrm{Dice}}+0.4\mathcal{L}_{\mathrm{BCE}}$ \\
Target radius / $\sigma$ (CARLA) & 7\,px / 2.5 \\
Target radius / $\sigma$ (RADIATE) & 7\,px / 7 \\
Simulation probe encoder & Frozen \\
RADIATE LR (enc./head) & $10^{-6}/10^{-3}$ \\
RADIATE epochs & 30 \\
\bottomrule
\end{tabular}
\end{table}

\begin{table*}[t]
\centering\small
\setlength{\tabcolsep}{5pt}
\caption{Simulation pretraining settings. All methods use four NVIDIA A100 GPUs with cross-device gathering of embeddings. RADICaL batch sizes count frames, each providing two augmented views.}
\label{tab:supp-current-pretrain-settings}
\begin{tabular}{@{}lcccc@{}}
\toprule
Setting & Vanilla CLIP & SG-CLIP & RankCLIP & RADICaL (adapted) \\
\midrule
Optimizer & AdamW & AdamW & AdamW & SGD, momentum 0.9 \\
Initial learning rate & $10^{-5}$ & $10^{-5}$ & $10^{-5}$ & $3\times10^{-3}$ \\
Weight decay & 0.05 & 0.05 & 0.05 & $10^{-4}$ \\
Epochs & 50 & 50 & 50 & 50 \\
Schedule & \multicolumn{3}{c}{Cosine, 5 warmup epochs from $10^{-7}$, minimum $10^{-6}$} & Cosine, no warmup \\
Batch per GPU / global & 160 / 640 & 160 / 640 & 160 / 640 & 56 / 224 \\
Temperature & 0.07 & 0.07 & 0.07 & 0.07 \\
Validation selection & Contrastive loss & SG-CLIP loss & CLIP loss & Two-view contrastive loss \\
\bottomrule
\end{tabular}
\end{table*}

\subsection{Data Partitions and Radar Input}
\label{sec:supp-current-data}

\paragraph{Simulation.}
All pretraining objectives use the same CARLA partition of 703,490 training, 87,922 validation and 87,971 test frames, obtained by a vehicle-count-stratified 80/10/10 split of scenes with up to 100 vehicles. Splits are disjoint at the frame level, and routes are shared across splits. The simulated log-amplitude range-angle heatmap is resized bicubically to $224\times224$, min-max scaled over the input batch, replicated across three channels and normalized with CLIP statistics.

\paragraph{RADIATE.}
Car, van, truck and bus are merged into one vehicle class. The sequence-disjoint partition contains 20,753 training, 3,630 validation and 8,614 test frames from 37, 6 and 17 sequences, respectively. Training and test follow the dataset's own assignments, and validation holds out complete recording groups from the training set. Only frames with at least one source annotation are indexed.

RADIATE intensities are converted to log-amplitude and mapped to $[0,1]$ using a calibration fitted on the training data. A centered $75\times75$\,m crop is resized bicubically to $224\times224$, and the same coordinate transform is applied to the annotations. Annotated boxes are retained when their centers lie inside the crop and their source size is at least 10 pixels. ViT models use three replicated channels with CLIP normalization~\cite{clip}; BBAVectors and U-Net use ImageNet normalization, and the rotated Faster R-CNN uses the input scaling of its pretrained detector.

\subsection{Radar-Language Pretraining}
\label{sec:supp-current-pretraining}

Each training frame is paired with one caption sampled at every iteration: a template caption with probability 0.25 and one of the LLM-generated variants otherwise (a 1:3 template-to-LLM ratio). Both forms describe the same vehicle-count record. Vanilla CLIP uses one-hot targets for the paired caption, while SG-CLIP uses the soft targets computed from the vehicle-count vectors of the samples in the gathered batch, with $\tau=0.07$ and $\alpha\in\{0.25,1,4\}$. Both encoders are trained jointly, starting from the OpenAI CLIP visual weights and a newly initialized text encoder. Table~\ref{tab:supp-current-pretrain-settings} lists the optimization settings for all pretraining objectives. A single validation-selected encoder per method is used for all downstream tasks and seeds.

\subsection{Downstream Training}
\label{sec:supp-downstream}

\paragraph{Simulation probes.}
Both simulation probes keep the pretrained radar encoder frozen. The captioning mapper and GPT-2 decoder are trained with teacher forcing on $\mathcal{L}_{\mathrm{caption}}$ using AdamW, a learning rate of $10^{-4}$ and a batch size of 32. The segmentation decoder is trained on four NVIDIA A10 GPUs with 256 frames per GPU (global batch 1,024) using the weighted Dice and BCE loss $\mathcal{L}_{\mathrm{center}}$ of the main paper.

\paragraph{RADIATE segmentation.}
Encoder and decoder are fine-tuned jointly for 30 epochs with $\mathcal{L}_{\mathrm{center}}=0.6\mathcal{L}_{\mathrm{Dice}}+0.4\mathcal{L}_{\mathrm{BCE}}$, using AdamW with weight decay 0.01 and constant encoder/head learning rates of $10^{-6}/10^{-3}$. U-Net uses an encoder learning rate of $10^{-4}$. Checkpoints are selected by validation IoU.

\paragraph{RADIATE detection.}
Encoder and detector are trained jointly with proposal and RoI classification and regression losses. Detection is trained with 5, 10, 20, 50 and 100\% of the labeled training frames. For each budget, every training sequence contributes one randomly placed contiguous block of frames; these blocks are nested across budgets and shared by all methods within a seed, while validation and test sets remain complete. Smaller budgets are trained for more epochs: the ViT detectors and BBAVectors use 270, 100, 50, 20 and 30 epochs for the five budgets, respectively. The ViT detectors select the checkpoint with the highest validation AP50.

\subsection{Evaluation Metrics}
\label{sec:supp-metrics}

\paragraph{Segmentation.}
Ground-truth masks place a normalized Gaussian at each vehicle center, so they encode where vehicles are rather than their physical extent. In simulation, centers come from CARLA actor positions projected into range-angle coordinates, with radius 7 pixels and $\sigma=2.5$. On RADIATE, centers are the annotated box centers, with radius 7 pixels and $\sigma=7$ on the native $432\times432$ crop, resized to the $224\times224$ output. Targets are binarized at 0.3 and predictions at the stated threshold. With $\mathrm{TP}$, $\mathrm{FP}$ and $\mathrm{FN}$ the true-positive, false-positive and false-negative pixels accumulated over the test set,
\[
\mathrm{IoU}=\frac{\mathrm{TP}}{\mathrm{TP}+\mathrm{FP}+\mathrm{FN}},\qquad
\mathrm{Dice}=\frac{2\,\mathrm{TP}}{2\,\mathrm{TP}+\mathrm{FP}+\mathrm{FN}}.
\]
Peak IoU is the maximum over a threshold sweep, and pixel AP is the area under the pixel-level precision-recall curve.

\paragraph{Captioning.}
The count-based captioning metrics are defined in Section~\ref{sec:supp-caption-simulation}.

\section{Baseline Implementations}
\label{sec:supp-baselines}

The representation baselines share the ViT-B/16 encoder, simulation data, input pipeline and downstream heads with SG-CLIP, so they differ only in the pretraining objective. The architecture baselines (BBAVectors and rotated Faster R-CNN for detection, U-Net for segmentation) use their own backbones and training recipes. For each annotation budget and seed, all detectors use the same training subset and the complete validation and test sets.

\subsection{RankCLIP}
RankCLIP~\cite{rank-clip} is trained on the same radar-caption pairs and adds cross-modal and intra-modal listwise ranking consistency to the symmetric CLIP objective. Radar-text, text-radar, radar-radar and text-text score matrices use the same temperature, and each pair of directed ListMLE terms is normalized by $32B$, where $B$ is the gathered batch size. For zero-indexed epoch $e$ of $E$, the ranking weight increases as
\begin{equation*}
 \lambda(e)=\min\!\left(2,\max\!\left(0,\frac{3e}{E-1}\right)\right).
\end{equation*}
Checkpoints are selected by the CLIP-only validation loss.

\subsection{RADICaL (Adapted)}
We adapt the radar-radar contrastive component of RADICaL~\cite{bootstrap-radar} to our ViT-B/16 encoder. Each frame yields two independently augmented views (random crop with side fraction 0.7, horizontal flip with probability 0.5) that are resized to the encoder input. A $512\!\rightarrow\!2048\!\rightarrow\!640$ projection head feeds a symmetric in-batch InfoNCE loss, with gradients through both views. This baseline uses no captions, count labels or camera supervision, and omits MIMO masking. It is an adaptation of RADICaL rather than a reproduction of the published recipe. The projection head is discarded for downstream tasks.

\subsection{BBAVectors}
BBAVectors~\cite{bbavectors2021} uses an ImageNet-initialized ResNet-18 with the original feature-combination, loss and decoding components. Its heads predict a center heatmap, box-boundary vectors and dimensions, sub-pixel center offsets and an orientation indicator. Retained oriented boxes are converted into these targets, and decoded vectors are converted back to rotated rectangles. The full network is trained.

\subsection{Rotated Faster R-CNN}
This baseline follows the RADIATE rotated Faster R-CNN configuration with a ResNet-50 FPN backbone~\cite{radiate,fasterrcnn2015,rrpn2018}, initialized from ImageNet weights, with one object class and anchor sizes of 8, 16, 32, 64 and 128 pixels. The stem and first residual stage are frozen; the remaining backbone, FPN, proposal network and RoI heads are trained.

\subsection{U-Net}
The segmentation reference is a U-Net~\cite{unet} with an ImageNet-initialized ResNet-34 encoder, bilinear decoding and a single sigmoid output. The radar channel is replicated three times and normalized with ImageNet statistics. It uses the same RADIATE partition, targets and loss as the ViT/PUP model.

\section{Attention Visualization}
\label{sec:supp-attention}

To inspect which parts of the radar input the pretrained encoder relies on before any downstream training, we visualize attention rollout~\cite{attention-flow2020} for the SG-CLIP encoder. For each of the final three transformer layers, the attention weights are averaged over the 12 heads, the identity matrix is added to account for the residual connection, and each row is renormalized. The resulting matrices are multiplied across layers, and the CLS-token row of the product assigns a weight to each of the 196 input patches. This $14\times14$ map is bilinearly upsampled to the $224\times224$ input and overlaid on the radar heatmap.

\begin{figure*}[t]
    \centering
    \includegraphics[width=\textwidth]{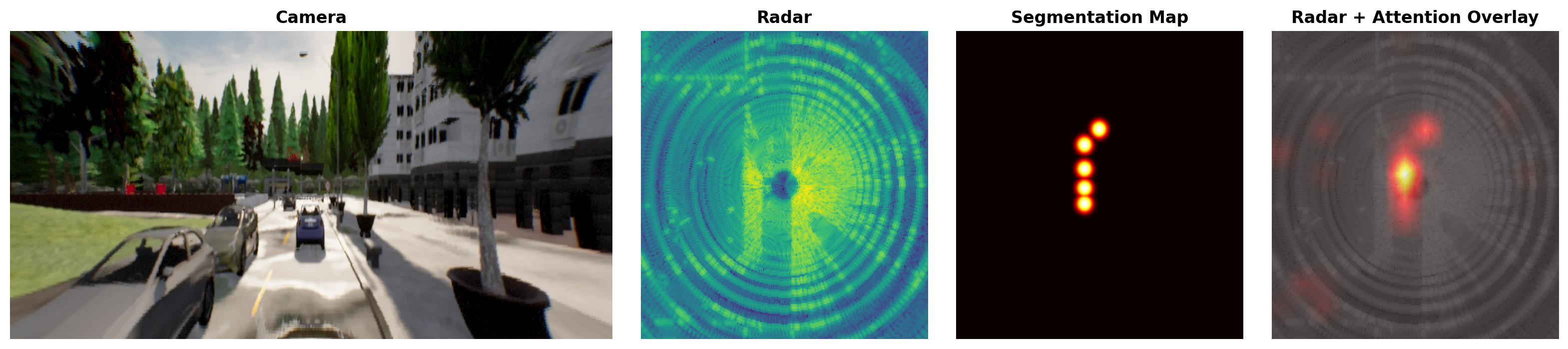}
    \caption{Attention rollout from the SG-CLIP pretrained ViT-B/16 encoder, aggregated over the final three transformer layers. From left to right: camera reference, radar input, ground-truth segmentation map, and attention overlaid on the radar heatmap. The encoder has not been trained on any downstream task.}
    \label{fig:supp-attention-prior}
\end{figure*}

During pretraining, the encoder sees only radar-caption pairs, without masks, boxes or other pixel-level targets. The CLS token is the embedding aligned with the caption, so its attention indicates which input regions contribute to the scene description. Any spatial selectivity in this map is therefore learned without localization labels.

In the scene of Figure~\ref{fig:supp-attention-prior}, attention concentrates on the column of vehicles ahead of the ego vehicle, with the strongest weight at near range, and on the farther vehicle to the right of this column. The attention panels in Figure~\ref{fig:supp-simulation-segmentation} show the same behavior on two further scenes: attention is strongest around occupied regions and is defined on the $14\times14$ patch grid.

Attention rollout is a qualitative visualization; the captioning, segmentation and detection results provide the quantitative evidence.

\section{Captioning Evaluation}
\label{sec:supp-caption-simulation}

\subsection{Count-Based Captioning Metrics}
\label{sec:supp-caption-spatial}

\begin{figure*}[t]
    \centering
    \begin{subfigure}[t]{0.48\textwidth}
        \centering
        \includegraphics[width=\linewidth]{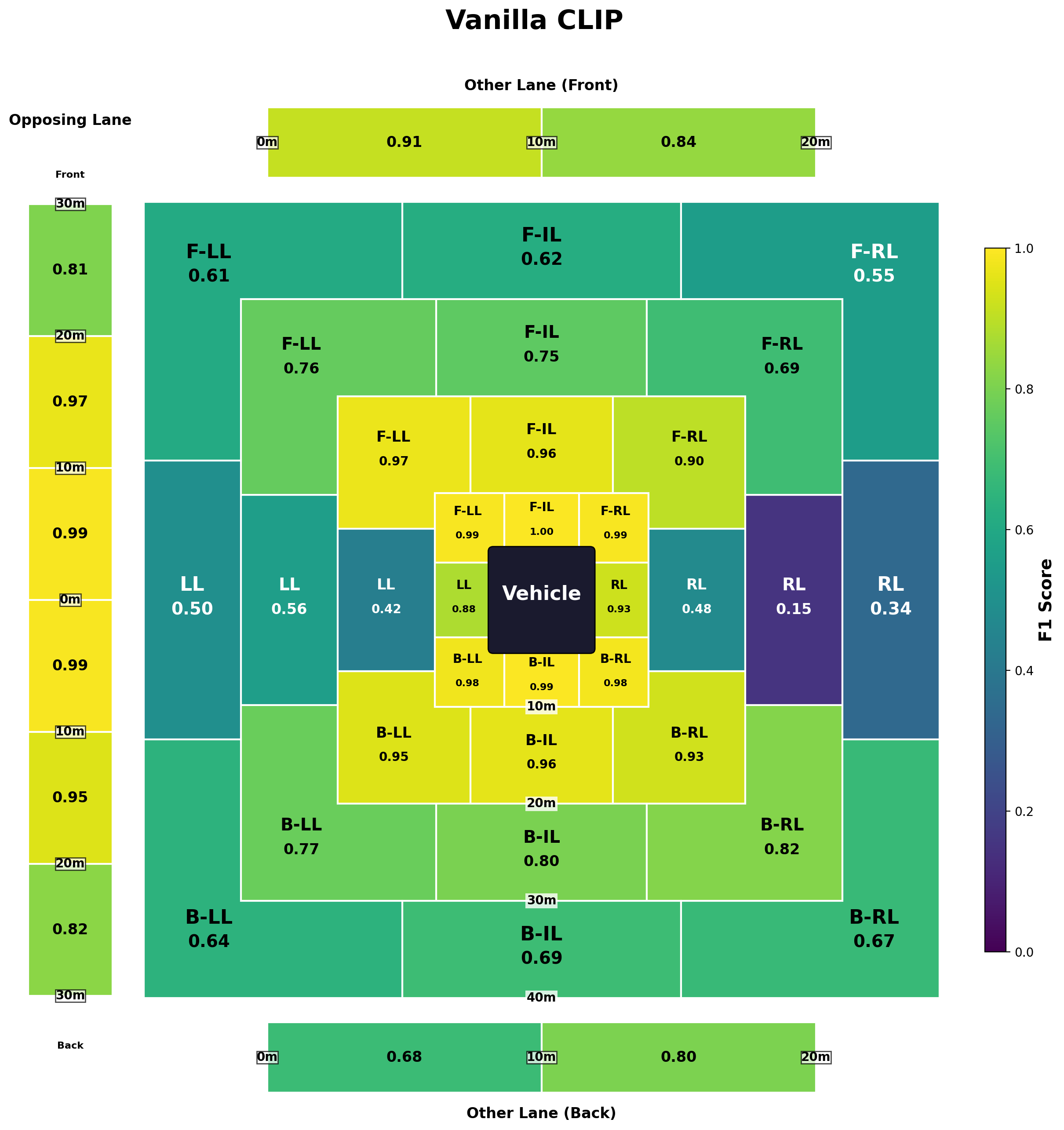}
        \caption{Vanilla CLIP.}
    \end{subfigure}\hfill
    \begin{subfigure}[t]{0.48\textwidth}
        \centering
        \includegraphics[width=\linewidth]{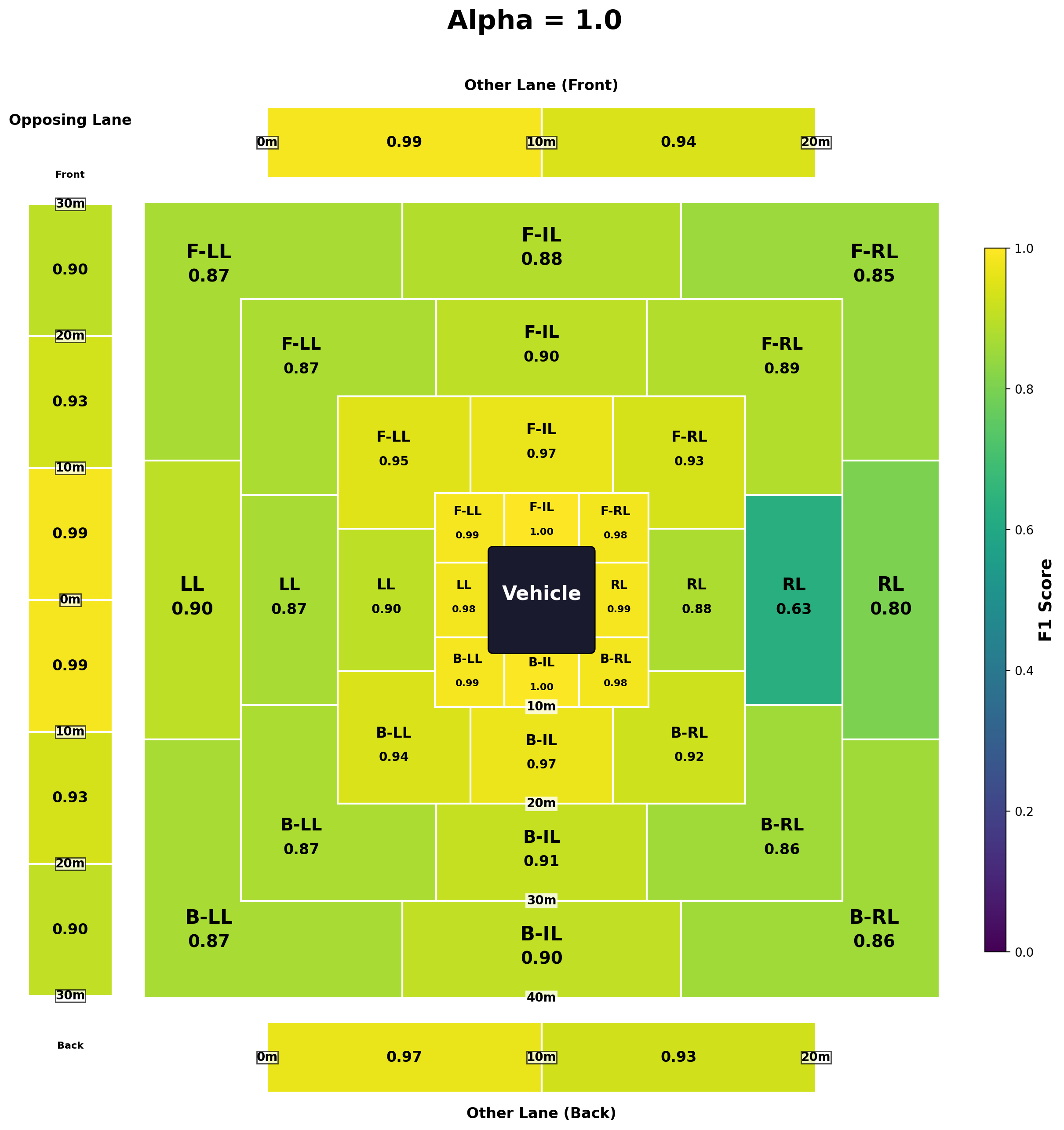}
        \caption{SG-CLIP ($\alpha=1$).}
    \end{subfigure}
    \caption{Per-cell captioning F1 by distance bin and relation on the simulation test set, with the radar encoder frozen. The ego vehicle is at the center and nested squares mark the 10\,m distance bins; side strips show the opposing and crossing lanes. Abbreviations: F/B front/back, LL/IL/RL left/in/right lane.}
    \label{fig:supp-caption-spatial-maps}
\end{figure*}

Lexical metrics do not test whether a caption places the correct number of vehicles in the correct region. We therefore score the vehicle counts recovered from each generated caption in every distance-relation cell $(b,s)$. For test scene $i$, let $y_{i,b,s}$ and $\hat y_{i,b,s}$ denote the annotated and recovered counts. Pooling over scenes gives
\begin{align*}
\mathrm{TP}_{b,s} &= \sum_i\min(\hat y_{i,b,s},y_{i,b,s}),\\
\mathrm{FP}_{b,s} &= \sum_i\max(0,\hat y_{i,b,s}-y_{i,b,s}),\\
\mathrm{FN}_{b,s} &= \sum_i\max(0,y_{i,b,s}-\hat y_{i,b,s}).
\end{align*}
True negatives are excluded because the descriptions enumerate vehicles that are present. The cell-wise metrics are
\begin{align*}
\mathrm{Precision}_{b,s} &= \frac{\mathrm{TP}_{b,s}}{\mathrm{TP}_{b,s}+\mathrm{FP}_{b,s}},\\
\mathrm{Recall}_{b,s} &= \frac{\mathrm{TP}_{b,s}}{\mathrm{TP}_{b,s}+\mathrm{FN}_{b,s}},\\
\mathrm{F1}_{b,s} &= \frac{2\mathrm{TP}_{b,s}}{2\mathrm{TP}_{b,s}+\mathrm{FP}_{b,s}+\mathrm{FN}_{b,s}}.
\end{align*}
The distance-bin scores in the main paper average $\mathrm{F1}_{b,s}$ over relations within each bin. Figure~\ref{fig:supp-caption-spatial-maps} shows the per-cell scores. Vanilla CLIP's lowest scores occur in the lateral cells beyond 10\,m and in the outermost bin. These are also where SG-CLIP gains most; for example, the right-side cell at 20-30\,m improves from 0.15 to 0.63 F1.

\subsection{Caption-to-Count Parser}
\label{sec:supp-caption-parser}

An LLM-assisted parser (GPT-5 nano) converts each generated caption into vehicle counts indexed by distance bin and relation, using the system prompt below; the schema's ellipses stand for the allowed keys. The parser sees only the generated caption, and ground-truth annotations are used only for scoring.

\begin{lstlisting}[language={},basicstyle=\ttfamily\scriptsize,breaklines=true,breakatwhitespace=false,columns=fullflexible,keepspaces=true]

A caption has been generated for a traffic scene based on some JSON data. The caption along with explanation of every key from the input JSON data has been provided. Your task is to regenerate the original JSON data as accurately as possible based on the caption and the explanation of every key. Use the caption to infer the values for each key in the JSON data. Here is the explanation of every key from the input JSON data:
0-10m: Objects within 0 to 10 meters from the ego vehicle. These are very close and may require immediate attention.
10-20m: Objects within 10 to 20 meters from the ego vehicle. These are close and should be monitored closely.
20-30m: Objects within 20 to 30 meters from the ego vehicle. These are at a moderate distance and should be noted.
30-40m: Objects within 30 to 40 meters from the ego vehicle. These are at a safe distance but still relevant.
left_lane_front_side: Vehicles in the left adjacent lane ahead of the ego vehicle.
left_side: Vehicles directly to the left side of the ego vehicle.
left_lane_back_side: Vehicles in the left adjacent lane behind the ego vehicle.
in_lane_front_side: Vehicles directly ahead of the ego vehicle in the same lane.
in_lane_back_side: Vehicles directly behind the ego vehicle in the same lane.
right_lane_front_side: Vehicles in the right adjacent lane ahead of the ego vehicle.
right_side: Vehicles directly to the right side of the ego vehicle.
right_lane_back_side: Vehicles in the right adjacent lane behind the ego vehicle.
opposing_lane_front: Vehicles in the opposing lane ahead of the ego vehicle.
opposing_lane_back: Vehicles in the opposing lane behind the ego vehicle.
other_lane_front: Vehicles in perpendicular or intersecting lanes ahead of the ego vehicle.
other_lane_back: Vehicles in perpendicular or intersecting lanes behind the ego vehicle.

The structure of the JSON data is as follows:
{
"0-10m": {
    "key_1": "value_1",
    "key_2": "value_2",
    ...
},
"10-20m": {
    "key_1": "value_1",
    "key_2": "value_2",
    ...
},
"20-30m": {
    "key_1": "value_1",
    "key_2": "value_2",
    ...
},
"30-40m": {
    "key_1": "value_1",
    "key_2": "value_2",
    ...
}
}
Note that value_1, value_2, etc should be in numerics (e.g., 0, 1, 2, ...). The keys inside each distance can be any of the following: "left_lane_front_side", "left_side", "left_lane_back_side", "in_lane_front_side", "in_lane_back_side", "right_lane_front_side", "right_side", "right_lane_back_side", "opposing_lane_front", "opposing_lane_back", "other_lane_front", "other_lane_back". If all keys in a distance bin is zero, do not output that distance bin. Only output the JSON data and nothing else.
\end{lstlisting}

The user message supplies the generated caption:

\begin{lstlisting}[language={},basicstyle=\ttfamily\scriptsize,breaklines=true,breakatwhitespace=false,columns=fullflexible,keepspaces=true]
Here is the generated caption:
{caption}
Using the caption and the explanation of every key from the original prompt, regenerate the JSON data as accurately as possible. Ensure that the structure of the JSON data matches the one provided in the system prompt. Also only output the JSON data and nothing else. Only output the key-value pairs that are non-zero for every distance bin.
\end{lstlisting}

\begin{table}[!h]
    \centering\footnotesize
    \setlength{\tabcolsep}{2.6pt}
        \caption{Simulation segmentation probe (\%). Precision, recall, IoU and Dice use prediction threshold 0.5; peak IoU is the maximum over the threshold sweep and pixel AP is the area under the pixel precision-recall curve.}
    \label{tab:supp-sim-center}
    \begin{tabular}{@{}lcccccc@{}}
        \toprule
        Method & Prec & Rec & IoU & Dice & \makecell{Peak\\IoU} & \makecell{Pixel\\AP} \\
        \midrule
        U-Net & 51.9 & \textbf{92.7} & 48.9 & 65.7 & 48.9 & 44.2 \\
        Vanilla CLIP & 82.0 & 69.9 & 60.6 & 75.5 & 61.5 & 52.2 \\
        \midrule
        SG-CLIP ($\alpha=0.25$) & 83.3 & 72.4 & 62.5 & 76.9 & 63.4 & 63.1 \\
        SG-CLIP ($\alpha=1$) & 82.6 & 71.7 & 62.3 & 76.8 & 63.1 & 62.8 \\
        SG-CLIP ($\alpha=4$) & \textbf{84.8} & 71.9 & \textbf{63.7} & \textbf{77.8} & \textbf{64.9} & \textbf{63.4} \\
        \bottomrule
    \end{tabular}
\end{table}

\section{Segmentation Results}
\label{sec:supp-center-transfer}

This section adds a tabular summary and examples for the simulation probe, followed by qualitative RADIATE results.

\subsection{Simulation Probe}
\label{sec:supp-center-sim}

\begin{figure*}[t]
    \centering
    \includegraphics[width=0.8\textwidth]{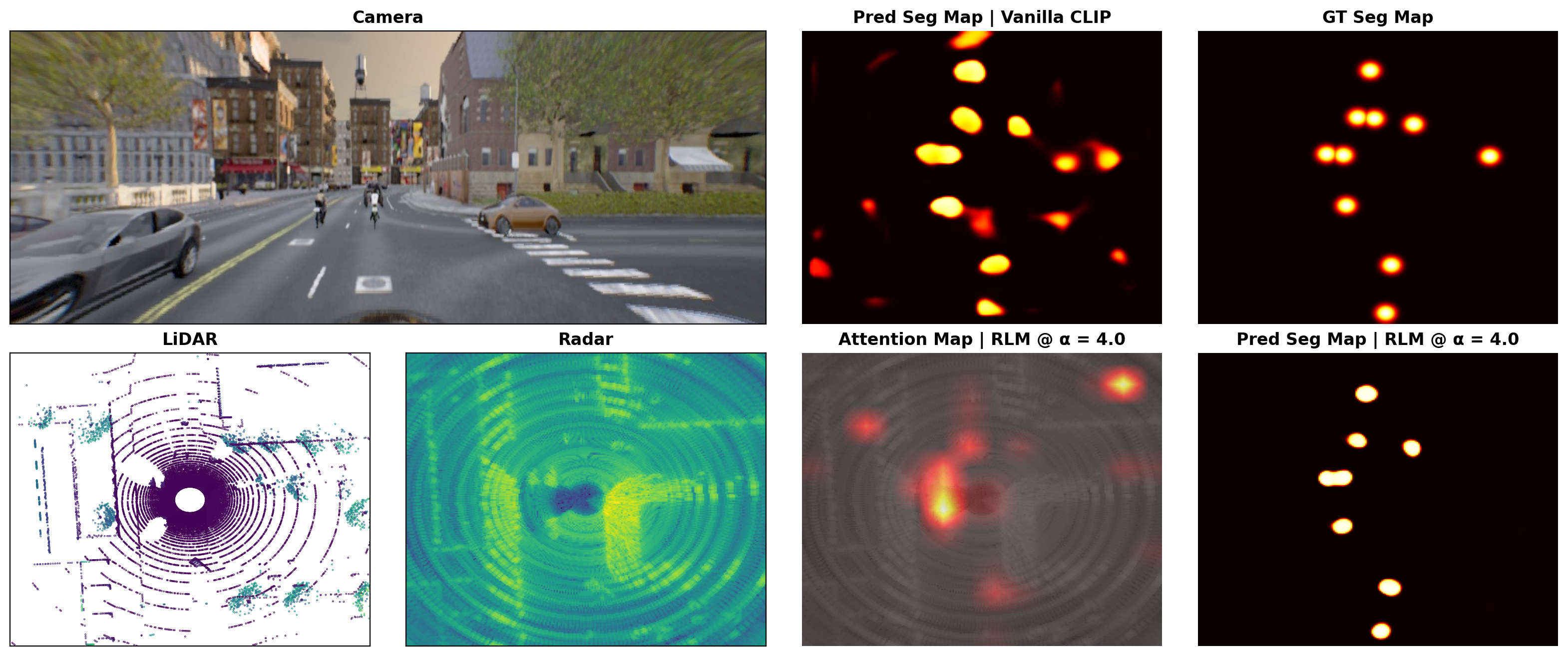}\par\vspace{6pt}
    \includegraphics[width=0.8\textwidth]{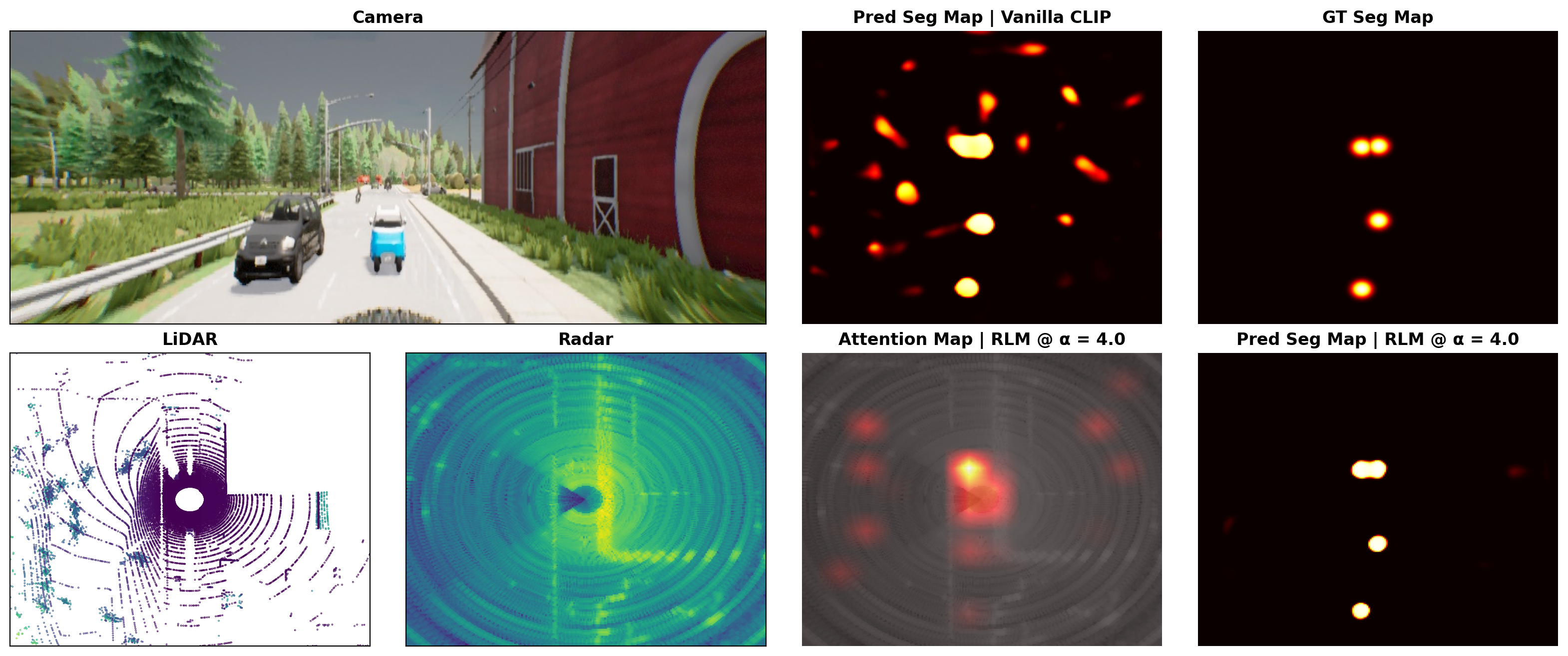}
    \caption{Simulation segmentation predictions with frozen encoders on two CARLA scenes. In each panel, the top row shows the camera view, the vanilla CLIP prediction and the ground truth. The bottom row shows LiDAR, radar input, SG-CLIP attention and the SG-CLIP prediction ($\alpha=4$, labeled RLM in the panels). Only the radar input is used for prediction.}
    \label{fig:supp-simulation-segmentation}
\end{figure*}

Table~\ref{tab:supp-sim-center} complements the simulation threshold curves reported in the main paper with fixed-threshold and threshold-free summaries. The ViT encoders are frozen and only the PUP decoder is trained, whereas U-Net is trained end to end as an architecture reference. All SG-CLIP variants improve over vanilla CLIP in precision, recall, IoU and Dice at threshold 0.5, as well as in peak IoU and pixel AP. The largest difference is in pixel AP (52.2 for vanilla CLIP and 62.8-63.4 for SG-CLIP), indicating better ranking of vehicle pixels across operating points.

Figure~\ref{fig:supp-simulation-segmentation} shows predictions on two CARLA test scenes, alongside the SG-CLIP attention map (Section~\ref{sec:supp-attention}). The camera and LiDAR views are shown only for context. In both scenes, vanilla CLIP produces many spurious responses away from the annotated vehicles, whereas SG-CLIP's responses are compact and coincide with the vehicles.

\subsection{RADIATE}
\label{sec:supp-center-radiate}

\begin{figure}[h]
    \centering
    \includegraphics[width=\columnwidth]{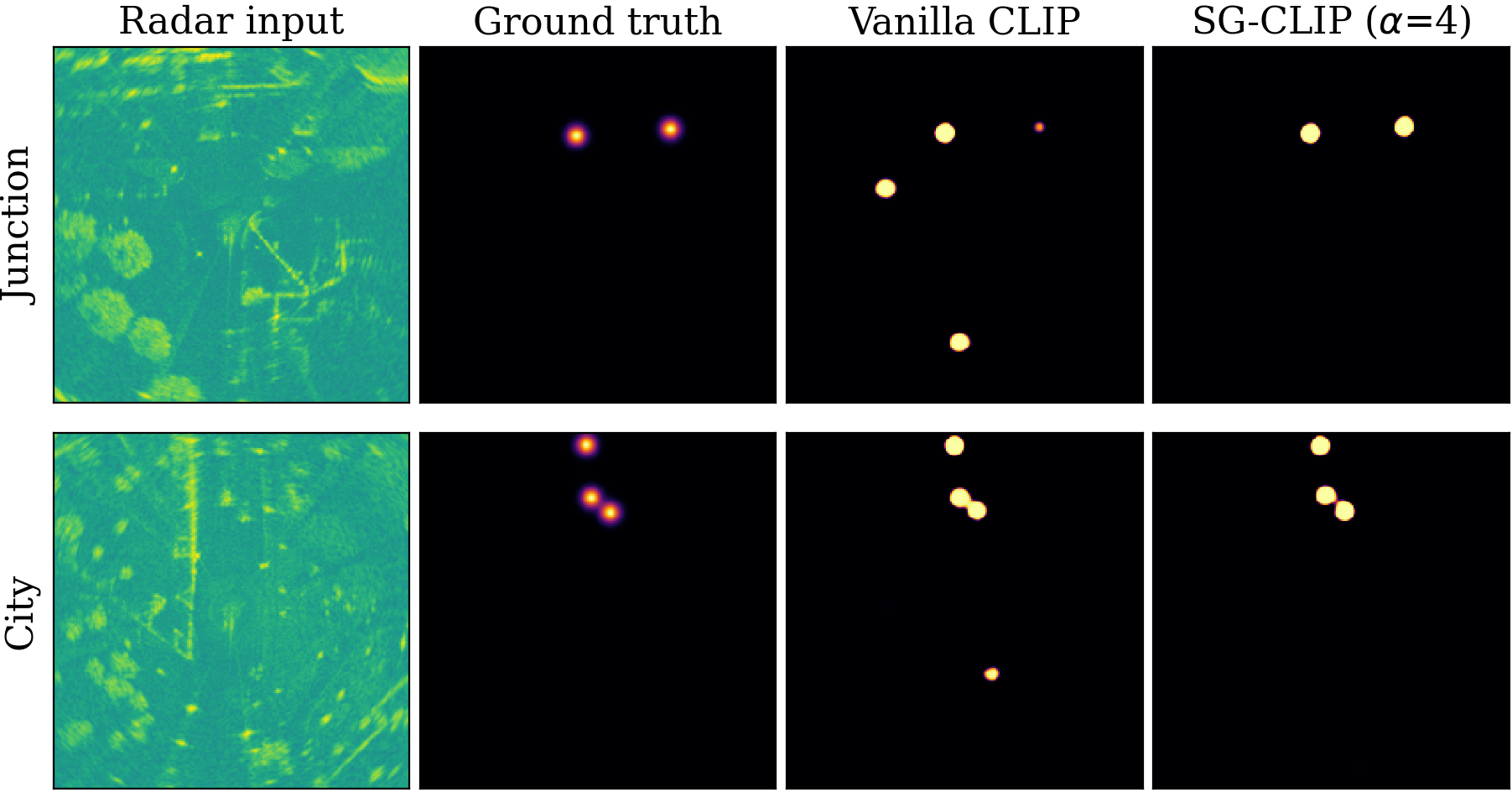}
    \caption{RADIATE segmentation predictions after fine-tuning with full labels, on a junction and a city test frame. Prediction maps show per-pixel probabilities; ground truth places a Gaussian at each annotated box center.}
    \label{fig:supp-radiate-center}
\end{figure}

Figure~\ref{fig:supp-radiate-center} shows fine-tuned segmentation predictions on two RADIATE test frames, from a junction and a city sequence. In both frames, vanilla CLIP produces responses away from the annotated vehicles, whereas SG-CLIP's responses coincide with the annotated vehicles.

\end{document}